\documentclass{aamas2014}

\usepackage{algorithm,algorithmic}
\usepackage{subfigure}
\def \imagewidth {4cm}

\begin{document}

\title{Finding Coordinated Paths for Multiple Holonomic Agents in 2-d Polygonal Environment}

\numberofauthors{3}

\author{
\alignauthor
Pavel Janovsk{\' y}\thanks{Equal contribution.}\\
	\affaddr{Agent Techology Center}\\
	\affaddr{FEL, CTU in Prague}\\
	\affaddr{Karlovo namesti 13, 121 35, Prague, Czech Republic}\\
	\email{{\normalsize pavel.janovsky@agents.fel.cvut.cz}}
\alignauthor
Michal \v{C}\'{a}p$^*$\\
	\affaddr{Agent Techology Center}\\
	\affaddr{FEL, CTU in Prague}\\
	\affaddr{Karlovo namesti 13, 121 35, Prague, Czech Republic}\\
	\email{{\normalsize michal.cap@agents.fel.cvut.cz}}
\alignauthor
Ji\v{r}\'{i} Vok\v{r}\'{i}nek\\
	\affaddr{Agent Techology Center}\\
	\affaddr{FEL, CTU in Prague}\\
	\affaddr{Karlovo namesti 13, 121 35, Prague, Czech Republic}\\
	\email{{\normalsize jiri.vokrinek@agents.fel.cvut.cz}}
}

\maketitle

\begin{abstract}
Avoiding collisions is one of the vital tasks for systems of autonomous mobile agents.
We focus on the problem of finding continuous coordinated paths
for multiple mobile disc agents in a 2-d environment with polygonal obstacles.
The problem is PSPACE-hard, with the state space growing exponentially in the number of agents. Therefore, the  state of the art methods include mainly reactive techniques and sampling-based iterative algorithms.

We compare the performance of a widely-used reactive method ORCA with three variants of a popular planning algorithm RRT* applied to multi-agent path planning and find that an algorithm combining reactive collision avoidance and RRT* planning, which we call ORCA-RRT* can be used to solve instances that are out of the reach of either of the techniques. We experimentally show that: 1) the reactive part of the algorithm can efficiently solve many multi-agent path finding problems involving large number of agents, for which RRT* algorithm is often unable to find a solution in limited time and 2) the planning component of the algorithm is able to solve many instances containing local minima, where reactive techniques typically fail.

%
%
\end{abstract}

\category{I.2.11}{ARTIFICIAL INTELLIGENCE}{Distributed Artificial Intelligence}[Intelligent agents, Multiagent systems]



\terms{Algorithms; Measurement; Performance; Experimentation}


\keywords{Path finding problem; multi-agent solver; planning; reactive technique}

\section{Introduction}
One of the fundamental tasks for a team of mobile agents is planning trajectories for the agents so as to avoid collisions between them. The problem arises in a variety of application domains such as in the teams of autonomous robots or in air traffic management systems.

The problem of finding coordinated paths for a group of agents is studied since 1980s. In theory it is well known that finding even non-optimal collision-free paths from defined start positions to goal positions for a group of objects in restricted 2-d space is PSPACE-hard \cite{hopcroft} and thus (unless P=PSPACE) there exist instances that cannot be solved in polynomial time.

Recent advancements in the field of multi-agent path planning include methods that solve restricted variant of the cooperative path finding problem, where point-like agents move using synchronous  discrete steps on a graph. For this variant of the problem there exist complete polynomial algorithms such as BIBOX \cite{surynek} or Push~\&~Rotate \cite{dewilde}, non-polynomial (but often effective) optimal algorithms \cite{standley} and anytime algorithms \cite{cap_2013_a}.  
However, the graph abstraction used in this problem formulation is not suitable for systems where the size of the agents cannot be neglected.

Research in trajectory planning for multiple agents has been made in the field of decoupled and centralized continuous planning domains. Decoupled approaches often offer better efficiency, but lose completeness. Frequently used decoupled technique is prioritized planning \cite{erdmann}. Here priorities are assigned to agents which specify the order in which their single-agent planning will take place. Each agent takes into account the plans of all agents with higher priority as moving obstacles. This method can be very efficient especially in uncluttered environments, but it is intrinsically incomplete. The method requires that the individual re-planning  takes place in the space-time. It is relatively straightforward to perform such planning with forward search algorithms such as A*, but it remains unclear how should be such a planning efficiently done in continuous space e.g. using some of the sampling-based algorithms.

Little attention has been devoted so far to the problem of finding coordinated trajectories for agents in continuous environments with polygonal obstacles. A straightforward solution is to construct the joint state space of all agents and search such a space using one of the sampling-based methods for continuous path planning. Alternatively, one can resort to a reactive collision avoidance techniques that proved to be very efficient in practice. The most prominent representatives of the two mentioned approaches are RRT* and ORCA.

In 2011 Karaman and Frazzoli \cite{karaman} published the RRT* algorithm, an anytime extension
of RRT (rapidly exploring random tree \cite{lavalle}) that is probabilistically complete and asymptotically optimal.
The algorithm is designed for continuous state spaces in which it can efficiently find a path from a given start state to a given goal
region by incrementally building a tree that is rooted at the start state and spans towards
randomly sampled states from some given state space. Once such a tree first reaches the
goal region, we can follow its edges backwards to obtain the first feasible path from
start state to the target region. With more and more samples being added to the tree, the solution incrementally improves.

Optimal reciprocal collision avoidance (ORCA) is one of the many methods based on velocity obstacle paradigm. It was proposed in 2011 by Jur van den Berg et al. \cite{vandenberg} and demonstrated to be capable of generating collision-free motion for a large number of agents in a cluttered workspace with local guarantees on optimality and safety.
Even though ORCA is a powerful tool for solving multi-agent collision avoidance, there are many instances of the problem that it is not able to solve due to the lack of cooperation and planning. An example of such an instance is any scenario where an agent has to make a long trip away from a narrow corridor to clear  the way to another agent.


The goal of this paper is to analyze the problem of planning collision-free continuous paths for multiple holonomic agents and evaluate the performance of the two approaches in 2-d polygonal environments.

We study the applicability of the RRT* algorithm for multi-agent path planning and compare it with ORCA -- one of the most widely used techniques for the problem, which has been successfully used in several software and hardware multi-agent implementations.
Moreover, we propose a novel variant of RRT* algorithm that is specifically suited for multi-agent path planning. The new ORCA-RRT* combines RRT* planning with reactive collision avoidance. Our experiments show that the combined algorithm is able to solve instances that were not solved by either of the techniques alone. Further, on many other instances the new algorithm provides higher quality solution that the other algorithms. 

The rest of the paper is structured as follows. In Section~2, we state the multi-agent path finding problem. Then, in Section~3 we  describe the multi-agent RRT* and introduce three variants of the algorithm that are applicable for multi-agent path planning in continous environments. The paper is concluded by a large-scale experimental analysis comparing the performance of the individual algorithms.


\section{Problem statement}\label{secProblemStatement}
We define the multi-agent path finding problem as follows. Consider $n$ agents operating in 2-dimensional Euclidean space with polygonal obstacles. The starting positions of agents are given as $n$-tuple $(s_1,\ldots,s_n)$, where $s_i$ is the starting position of $i$-th agent. The $n$-tuple $(d_1,\ldots,d_n)$ gives the agents' destinations. We assume that the agents have disc-shaped bodies, where the radius of agent $i$ is denoted as $r_i$.
The final trajectory of $i$-th agent $\pi_i (t)$ is a mapping $\mathbb{R} \rightarrow \mathbb{R}^2$ of time $t$ to 2~dimensional Euclidean space of the agent, where $\pi_i (0) = s_{i}$. The time $t_{i}^{d}$ is the minimal time after which the $i$-th agent remains at its destination,
\begin{equation*}\label{tiMax}
t_{i}^{d} = \min(t_{i}| \forall t \in \langle{t_{i}, \infty}) : \pi_{i}(t)=d_{i}).
\end{equation*}

Let $O\subset \mathbb{R}^2$ denote the regions of the space occupied  by the obstacles. Then, the collision-free property of the set of multi-agent trajectories can be defined as
\begin{eqnarray*}\label{collisionFree}
\lefteqn{CF(\{\pi_1,\ldots,\pi_n\}, O) = true \text{ iff}} \\
  &&\forall i,j,t, i \neq j: dist(\pi_i (t),\pi_j (t)) > r_i + r_j \nonumber \\
  &&and  \nonumber \\
  &&\text{ }\forall i,t: D(\pi_i (t),r_i) \cap O = \emptyset\nonumber,
\end{eqnarray*}
where $dist(x,y)$ is Euclidean distance in 2 dimensional space and $D(\pi_i (t),r_i)$ denotes a disc of radius $r_i$ centered at $\pi_i (t)$. The task is to find $n$-tuple $(\pi_1,\ldots,\pi_n)$ such that the agents never collide and the sum of times agents spend on the path to their final destinations is minimal. The problem statement is defined as follows:
\begin{eqnarray*}\label{problemStatement}
\lefteqn{\text{Find }(\pi_1,\ldots,\pi_n) \text{ s. t.}} \\
  &&CF(\{\pi_1,\ldots,\pi_n\}, O) = true \nonumber \\
  &&\sum\limits_{i=1}^n t_{i}^{d} \text{ is minimal.}\nonumber
\end{eqnarray*}

\section{RRT* for multi-agent path finding}\label{secRRT}
The RRT* algorithm was originally designed for single-agent motion planning, however, it can be adapted to solve multi-agent path planning problems. In this section we will discuss how can be the RRT* algorithm leveraged in a multi-agent setting.



We let the RRT* algorithm operate in the joint state space $J = C_1 \times \ldots \times C_n$, where $C_i \subseteq \mathbb{R}^2$ is the state space of the $i$-th agent. The initial state in the joint state space is $\mathbf{x}_{init}=(s_1,\ldots,s_n)$, goal state is $\mathbf{x}_{goal}=(d_1,\ldots,d_n)$ and any other sampled state is also $n$-tuple containing positions of all agents. The algorithm searches for a path $\mathbf{p}: [0,1] \rightarrow J$ from $\mathbf{x}_{init}$ to $\mathbf{x}_{goal}$, which can be then decomposed into a set trajectories $\{\pi_i\}$ for each individual agent $i$. The pseudocode of RRT* algorithm for multi-agent path planning is exposed in Algorithm~\ref{alg:RRTStar}.

The distance metric used in the algorithm is the sum of Euclidean distances of all $n$ agents:

\begin{equation*}
dist(\mathbf{x},\mathbf{y}) = \sum \limits_{i=1}^n dist_{E}(x_i,y_i),
\end{equation*}
where $x,y \in J$ and $x_{i},y_{i} \in C_i$.

Now, we provide definitions of the following RRT* primitive procedures:

{\it Nearest Neighbor:} Given a graph $T = (V, E)$ and a state $\mathbf{x} \in J$, the function
$Nearest(T,\mathbf{x})$ returns a vertex $\mathbf{v} \in V$ that is the closest to state $\mathbf{x}$ in terms of the distance metric $dist(\cdot,\cdot)$.

{\it Near Vertices:} Given a graph $T = (V, E)$, a state $\mathbf{x} \in J$ and the numbers $n, m \in \mathbb{N}$,
the function $Near(T, \mathbf{x}, n)$ returns a set $\{\mathbf{y} : \mathbf{y} \in V \wedge dist(\mathbf{x},\mathbf{y}) < r_n\}$,
 where $r_n = \gamma (\log n / n)^{1/d}$, $\gamma$ is a constant and $d=2n$ is the dimension of the space $J$.

{\it Local Steering Procedure:}
Given two states $\mathbf{x}$ and $\mathbf{y}$, a domain-specific local steering procedure $Steer(\mathbf{x},\mathbf{y})$ returns true if the steering procedure is able to connect the state $\mathbf{x}$ to the state $\mathbf{y}$. In the context of multi-agent path finding, the steering procedure seeks for a set of paths $\{p_i\}$ such that $p_i(0)=x_i$ and $p_i(1)=y_i$ for each agent $i$ and all paths are mutually collision-free. We consider three different methods for local steering which are discussed in detail in Section \ref{secExtensions}.


{\it Find best parent:} Given a graph $T = (V, E)$, a set $X \subseteq V$ and a state $x \in J$, the function $FindBestParent(T,X,x)$ returns the best parent vertex $v \in X$ for state $x$, i.e. the vertex that yields the lowest cost of path $\mathbf{x}_{init} \rightarrow \mathbf{x}$.

{\it Rewire:} Given a graph $T = (V, E)$, a set $X \subseteq V$ and a state $x \in J$ the function $Rewire(T,X,x)$ examines all vertices from $X$ to see whether their cost can be improved by going through the new vertex $x$. If there are any such vertices, the tree is
rewired so that these vertices become children of $x$, which allows them to
improve the cost of their path from the initial state $\mathbf{x}_{init}$.

The main loop of the multi-agent RRT* algorithm (see Algorithm~\ref{alg:RRTStar}) works as follows: Until interrupted,
the algorithm samples states from the joint state space $J$ (line~\ref{algSample}). Each such a random sample is used in an attempt to
extend the tree. First, the nearest vertex from the tree $T$ to the random sample $\mathbf{s}$ is determined (line~\ref{algNearest}).
Then, the algorithm attempts to connect the vertex $\mathbf{x}$ to the random sample $\mathbf{s}$ using the local steering procedure (line~\ref{algSteer}). If the two points can be connected using the local steering procedure the algorithm proceeds to its optimizing part, otherwise it tries again with another sample. Then a set of vertices $X_{near}$ that are within a specified distance to the $\mathbf{x}$ is determined (line~\ref{algNear}). The $X_{near}$ set serves two purposes: Firstly it is used to find the best parent for the new vertex $\mathbf{x}$ (line~\ref{algBestParent}). Secondly, after the new vertex $\mathbf{x}$ is added to the tree, the vertices from $X_{near} $ are rewired if it improves their cost (line~\ref{algRewire}).

Once the goal state\footnote{Here we assume that the point $\mathbf{x}_{goal}$ is sampled with a certain non-zero probability (e.g. 1\%) to overcome the need to define the goal as a region instead of a point.} is successfuly added to the tree, one can follow the links backwards to obtain a valid path from the state $\mathbf{x}_{init}$ to the state $\mathbf{x}_{goal}$.  However, even after the first solution has been returned, the algorithm does not stop iterating. Instead, the algorithm continously extends and rewires the tree, which leads to incremental discovery of further higher-quality paths.

\begin{algorithm}[H]
\begin{algorithmic}[1]
\algsetup{linenosize=\small}
\STATE $V \leftarrow \{\mathbf{x}_{init}\} ; E \leftarrow \emptyset ;$
\WHILE{not interrupted}
\STATE $T = (V, E);$
\STATE $\mathbf{s} \leftarrow Sample();$  \label{algSample}
\STATE $\mathbf{x} \leftarrow Nearest(T, \mathbf{s});$  \label{algNearest} 	
\IF{ $Steer(\mathbf{x}, \mathbf{s})$ } \label{algLineSteering}  \label{algSteer}  	
	\STATE $X_{near} \leftarrow Near(T, \mathbf{s}, |V|);$   \label{algNear}
	\STATE $\mathbf{x}_p \leftarrow FindBestParent(T, X_{near}, \mathbf{s});$ \label{algBestParent}
	\STATE $V \leftarrow V \cup \{\mathbf{x}\}; E \leftarrow E \cup \{(\mathbf{x}_p, \mathbf{x})\};$
	\STATE $Rewire(T, X_{near}, \mathbf{s});$ \label{algRewire}
\ENDIF

\ENDWHILE
\end{algorithmic}
\caption{Multi-agent RRT*}
\label{alg:RRTStar}
\end{algorithm}

\subsection{Multi-agent Extensions}\label{secExtensions}

The crucial component influencing the performance of the RRT* algorithm is the local steering procedure.  We consider a multi-agent steering function of the following form:

\begin{eqnarray*}
\lefteqn{Steer(\mathbf{x}, \mathbf{y}) =}\\
 &&\left\{
  \begin{array}{l l}
    true         & \text{if } \exists\: \{\pi_i\} \text{ s.t. } \{\pi_i\} = E(\mathbf{x}, \mathbf{y},O)\\
    		&  \quad \text{ and } CF(\{\pi_i\},O)\\
    		&  \quad \text{ and }  \exists t \; \forall i \; \pi_i(t)=y_i\\
  false         &  \text{otherwise}  \\
  \end{array} \right. ,
\end{eqnarray*}
where $\mathbf{x},\mathbf{y} \in J$. The function $E(\mathbf{x}, \mathbf{y},O): J \times J \times \mathcal{P}(\mathbb{R}^2)  \rightarrow \Pi_1 \times \ldots \times \Pi_n$ is an extension function that returns a set of paths $\{\pi_i\}$, where $\pi_i$ is a path for agent $i$ such that $\pi_i(0)=x_i$ and $\exists t_d \text{ s.t. } \forall t \geq t_d :\; \pi_i(t)=y_i$. Note that the steering procedure returns true only if the trajectories generated by the extension function are 1) avoiding all static obstacles and 2) the trajectories are mutually collision-free.

We study three methods that can serve as a valid extension function for the multi-agent steering procedure: 1) {\em Line} extension connects individual agents using straight lines, 2) {\em VisibilityGraph} extension connects individual agents using the optimal single-agent path, and 3) ORCA extension generates joint path by simulating ORCA algorithm, where each agent tries to follow its single-agent optimal path.

\subsubsection{Line Extension}

The first considered extension method is a relatively straightforward extension of classical single-agent RRT* planner as exposed in~\cite{karaman}, which connects two states using straight line paths. In a multi-agent setting, we connect each of the agents by a straight path:
$$
E_{Line}(\mathbf{x}, \mathbf{y},O) = \{line(x_i,y_i)\},
$$
where the $line(x,y): \mathbb{R}^2\times\mathbb{R}^2 \rightarrow \Pi$ is a function defined as
$$
line(x_i,y_i)=\pi_i(t) = \left\{
  \begin{array}{l l}
    x_i + v_i\cdot t \frac{y_i-x_i}{|y_i-x_i|}	& \text{for } t < \frac{|y_i-x_i|}{v_i}\nonumber\\
    y_i		& 	 otherwise\\
  \end{array} \right. ,
$$
where $v_i$ denotes the maximal speed of the $i$-th agent. Observe that such a trajectory prescribes that the agent $i$ should move to the point $y_i$ along the straight line at maximum speed; when the point $y_i$ is reached, the agent stays there.
Note that it can happen that a solution returned by this method is rejected by the steering procedure if a) any of the trajectories intersects some of the obstacles or b) some of the trajectories are in mutual collision.

\subsubsection{Visibility Graph Extension}
Another considered method uses the single-agent optimal paths (i.e. optimally avoids obstacles, but ignores interactions among the agents) between the points $x_i$ and $y_i$. In 2-d polygonal environments, one can efficiently find such a~path by constructing a visibility graph and subsequently by finding the shortest path in such a graph. Let the shortest path in the visibility graph of $i$-th agent be represented as a~sequence of edges $(e_{i1},\ldots,e_{im_{i}})$. The resulting trajectory of agent $i$ will follow this path. The {\it Visibility Graph} extension function can be formally defined as
$$
E_{VG}(\mathbf{x}, \mathbf{y},O) = \{segments(x_i,y_i)\},
$$
where $segments(x_i,y_i): \mathbb{R}^2\times\mathbb{R}^2 \rightarrow \Pi$ is a function defined as

\begin{eqnarray*}
\lefteqn{segments(\mathbf{x_i},\mathbf{y_i})=\pi_i(t)}\\
&&=\left\{
  \begin{array}{l}
   source(e_{ij}) + v_i\cdot t \frac{e_{ij}}{|e_{ij}|}	\nonumber\\
    \quad \quad \text{for } t \in (\frac{\sum\limits_{k=1}^{j-1}|e_{ik}|}{v_{i}}, \frac{\sum\limits_{k=1}^{j}|e_{ik}|}{v_{i}}), j \in \langle 1,m_{i}\rangle\nonumber\\
    y_i \quad otherwise\\
  \end{array} \right. ,
\end{eqnarray*}
where $source(e)$ function returns the source vertex of the edge $e$. Note that the trajectories returned by this method are guaranteed to be obstacle avoiding, therefore, the steering procedure may only reject them if the resulting trajectories are in mutual collision.

\subsubsection{ORCA Extension}
The last considered extension method generates trajectories by following optimal single-agent paths together with  Optimal Reciprocal Collision Avoidance (ORCA) technique, which is used for reactive collision avoidance between the agents. ORCA is a decentralized reactive collision avoidance method based on the velocity-obstacle paradigm that performs optimization in the space of velocity vectors for each agent. During ORCA, each agent creates a velocity obstacle for every other agent based on their currently observed position and velocity. Each such a velocity obstacle cuts out a half-plane in the space of possible velocities for the agent. Given the agent's desired velocity and the constraints induced by the other agents, a linear program with $n-1$ constraints is constructed and solved to obtain the optimal velocity vector the agent should follow in the next time step \cite{vandenberg}.

To provide the desired velocity vector in each time step, we use the same visibility graph based navigation as in the visibility graph extension method. At each timestep, the agent finds the optimal path from its current position to its destination on the visibility graph and sets the desired velocity vector to point at this direction.


The ORCA extension function is defined as
\begin{equation*}
E_{ORCA}(\mathbf{x}, \mathbf{y},O) =\{\pi_i\}
\end{equation*}
where $\pi_i$ is a trajectory of $i$-th agent obtained by simulating the ORCA method with $\mathbf{x}$ as start positions and $\mathbf{y}$ as goal positions.

The trajectories returned by ORCA are guaranteed to be obstacle avoiding. They are also guaranteed to be mutually collision-free. However, due to the reactive nature of the algorithm, the method is not guaranteed to find a solution if a solution exists. Since some of the executions may end up in infinite loops or deadlocks, it is not uncommon to see the agents being stuck at one point and never reach their designated target position. Therefore one has to bound the maximum number of timestep each ORCA simulation is allowed to perform.

Our hypothesis is that there is a  significant number of problem instances that cannot be solved by ORCA alone, but that could be efficiently solved by multi-agent RRT* with ORCA extensions.

\begin{figure}[ht]
\centering
\subfigure[two disk shaped agents exchange their positions]{\includegraphics[width = 3in]{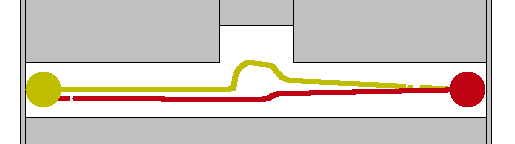}}\\
\subfigure[two teams of disk shaped agents exchange their positions]{\includegraphics[width = 3in]{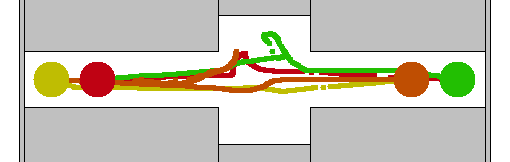}}
\caption{Example of instances solved by ORCA-RRT* that was not solved by ORCA (lines show trajectories $\pi_{i}(t)$)}
\label{fig:orca-rrt-instances}
\end{figure}

An example of an artificial instance that was not solved by ORCA is in Figure~\ref{fig:orca-rrt-instances}. In such a scenario, the reactive technique tryies to resolve the conflict by letting both agents to go  back in the corridor, but at some point it decides to return back to its desired directon, resulting in a deadlock situation.

\newpage

\begin{figure*}
\centering
\subfigure[Empty environment]{\includegraphics[width=3.9cm]{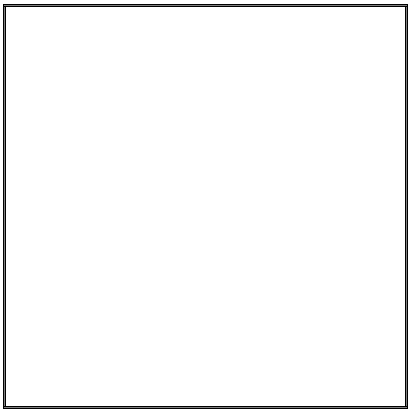}}
\subfigure[Door environment]{\includegraphics[width=3.9cm]{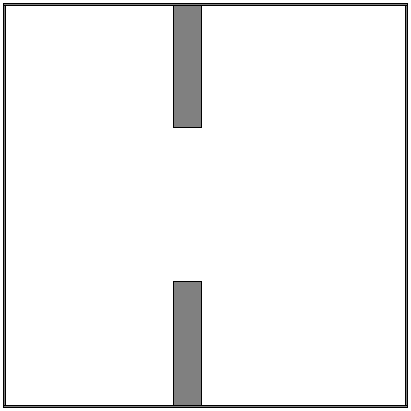}}
\subfigure[Cross environment]{\includegraphics[width=3.9cm]{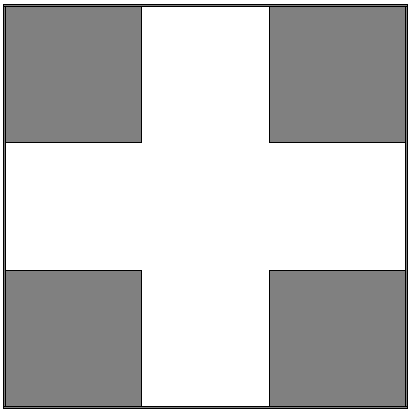}}
\subfigure[Maze environment]{\includegraphics[width=3.9cm]{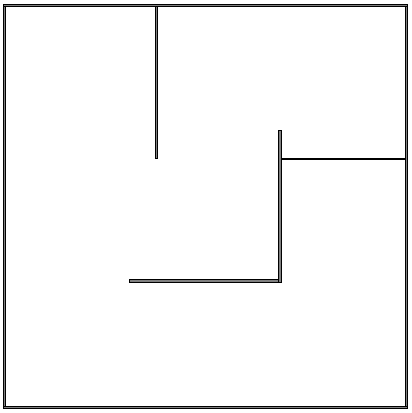}}
\vspace{-3mm}
\caption{Problem environments}
\label{figEnvironments}
\vspace{-3mm}
\end{figure*}

\section{Experimental analysis}\label{secExpAnalysis}
In this section we experimentally evaluate features of proposed extension of multi-agent RRT* algorithm. We compare three RRT* based algorithms - Line RRT*, Visibility Graph RRT* and ORCA RRT* with reactive ORCA. The experiments have been performed on four types of 2-d testing environments -- empty, door, cross and maze environments (see Figure~\ref{figEnvironments}). All environments are constrained by a fixed boundary having 1000x1000 dimension. The metrics for comparison has been focused on success rate of the algorithms and the quality of the solution for various environment settings. The measured parameters are:
\begin{itemize}
\item {\bf idealistic solution cost} is the cost of a solution for which the $CF$ function in Equation \ref{collisionFree} is relaxed in a~way that it omits its first constraint i.e. permits collisions between agents. The idealistic (i.e. lower bound) solution cost is defined as
    \begin{center}
     $t_{ideal} = \sum\limits_{i=1}^n t_{i}^{d} .$
    \end{center}
     The goal arrival time $t_i^d$ is obtained by computing a~single-agent optimal path for each agent using the visibility graph.
\item {\bf suboptimality} measure gives indicative quality ratio by comparison of the given solution to the lower bound of the solution provided by idealistic solution cost. It shows how many times is the given solution worse than the idealistic solution. It is defined as
     \begin{center}
     $suboptimality = \frac{\sum\limits_{i=1}^n t_{i}^{d}}{t_{ideal}}$.
     \end{center}
\item {\bf success rate} shows the percentage of scenarios solved by the algorithms. The success rate depends on suboptimality in a way that a given solution is successful only if its suboptimality is lower than a defined \emph{threshold}. If the algorithm does not provide any solution within time frame defined by experiment setting, it is automatically considered unsuccessful.
\end{itemize}


\subsection{Benchmark set}
The experiment has been performed on a benchmark set composed of four environments as depicted in Figure~\ref{figEnvironments} with the number of agents varying from 2 to 10 and the agent body radii ranging from 50 to 100. For each combination of environment, number of agents and agent radius, the benchmark set contains 10 different settings of agents' start and goal positions, altogether 2160 benchmark scenarios. Since RRT* is a stochastic algorithm, we run the algorithm five times with different random seeds for each problem instance. Altogether, the presented experimental results are based on 34560 runs.

The benchmark set was created by an algorithm that guarantees that for each problem instance there is exactly one collision cluster i.e. the path finding problem cannot be divided into mutually non-colliding groups of agents, which would be easier to solve. This is guaranteed by adding agent's random start and goal positions iteratively only when a collision occurs between agent's shortest path from start to goal and any other agent's shortest path. The procedure that creates a problem instance is given in Algorithm~\ref{alg:Creator}.

\begin{algorithm}
\begin{algorithmic}[1]
\algsetup{linenosize=\small}
\FOR{i in 1:numberOfAgents}
\STATE colliding = false
\WHILE{not colliding}
\STATE start = randomSample
\STATE goal = randomSample
\STATE path = findShortestPath(start, goal, obstacles)
\IF{path exists}
\STATE colliding = findCollision(path, allPaths)
\ENDIF
\ENDWHILE
\STATE allPaths.add(path)
\STATE starts.add(start)
\STATE goals.add(goal)
\ENDFOR
\end{algorithmic}
\caption{Create problem instance}
\label{alg:Creator}
\end{algorithm}

\begin{figure}
\centering
\subfigure[]{\includegraphics[width=3.8cm]{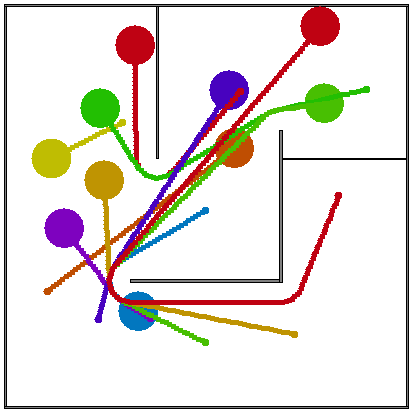}}
\subfigure[]{\includegraphics[width=3.8cm]{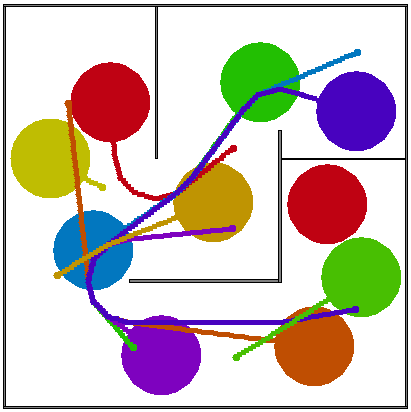}}
\vspace{-3mm}
\caption{Example of maze environment -- 10 agents with radii a) 50, b) 100 with corresponding single agent optimal paths.}
\label{figInt}
\end{figure}

An example of one such generated problem instance is depicted in Figure~\ref{figInt}. It shows the maze environment with 10 agents of minimal radus $r=50$ and maximal radius $r=100$ and the corresponding optimal single-agent trajectories obtained by running the A* algorithm on a visibility graph.


In the next sections we will discuss the results of experimental evaluation from the perspective of the instance set coverage/success rate of the algorithms and the solution quality/suboptimality.

\begin{figure*}[p]
\centering

\subfigure[Empty environment]{\includegraphics[width=\imagewidth]{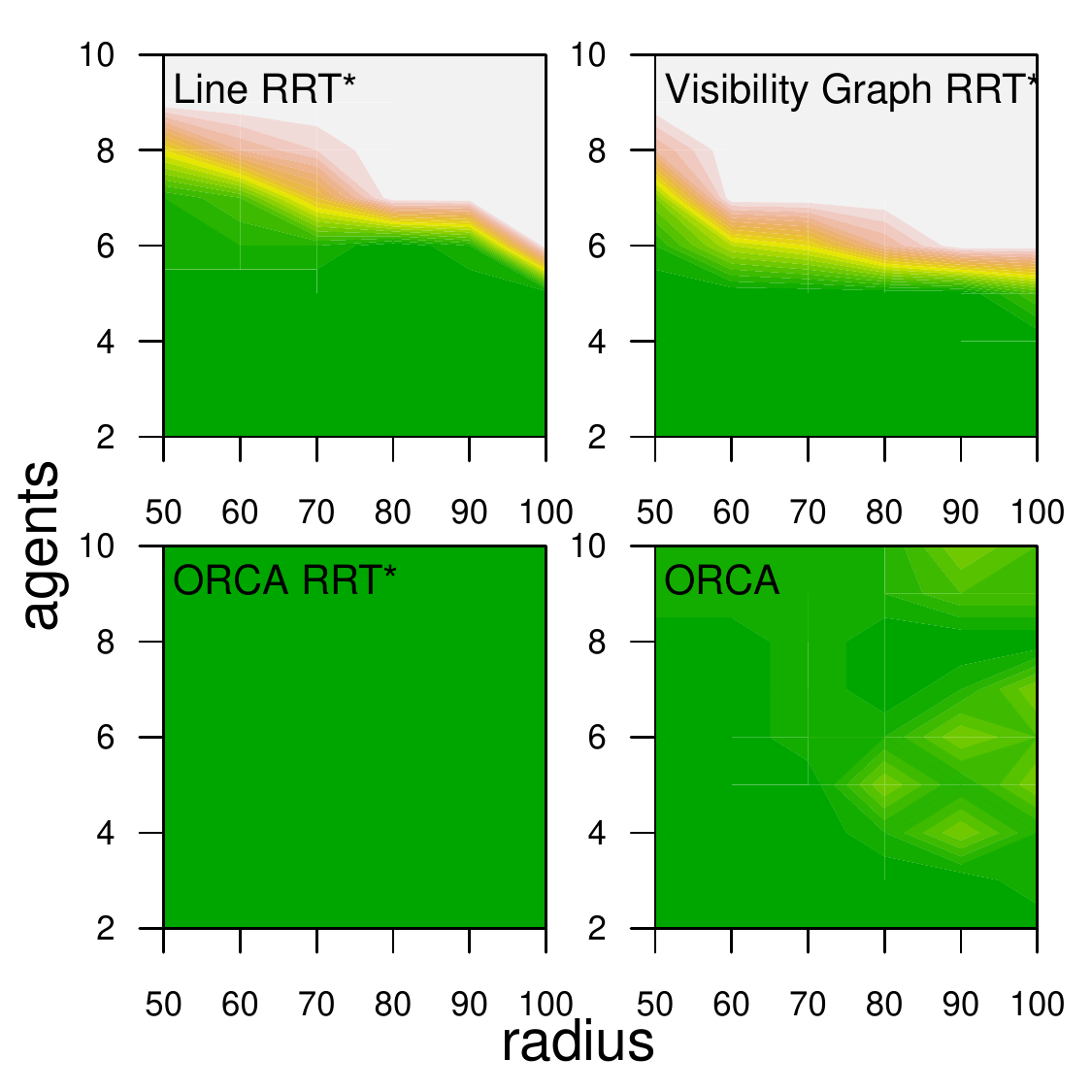}\label{figSuccessrateFree}}
\subfigure[Door environment]{\includegraphics[width=\imagewidth]{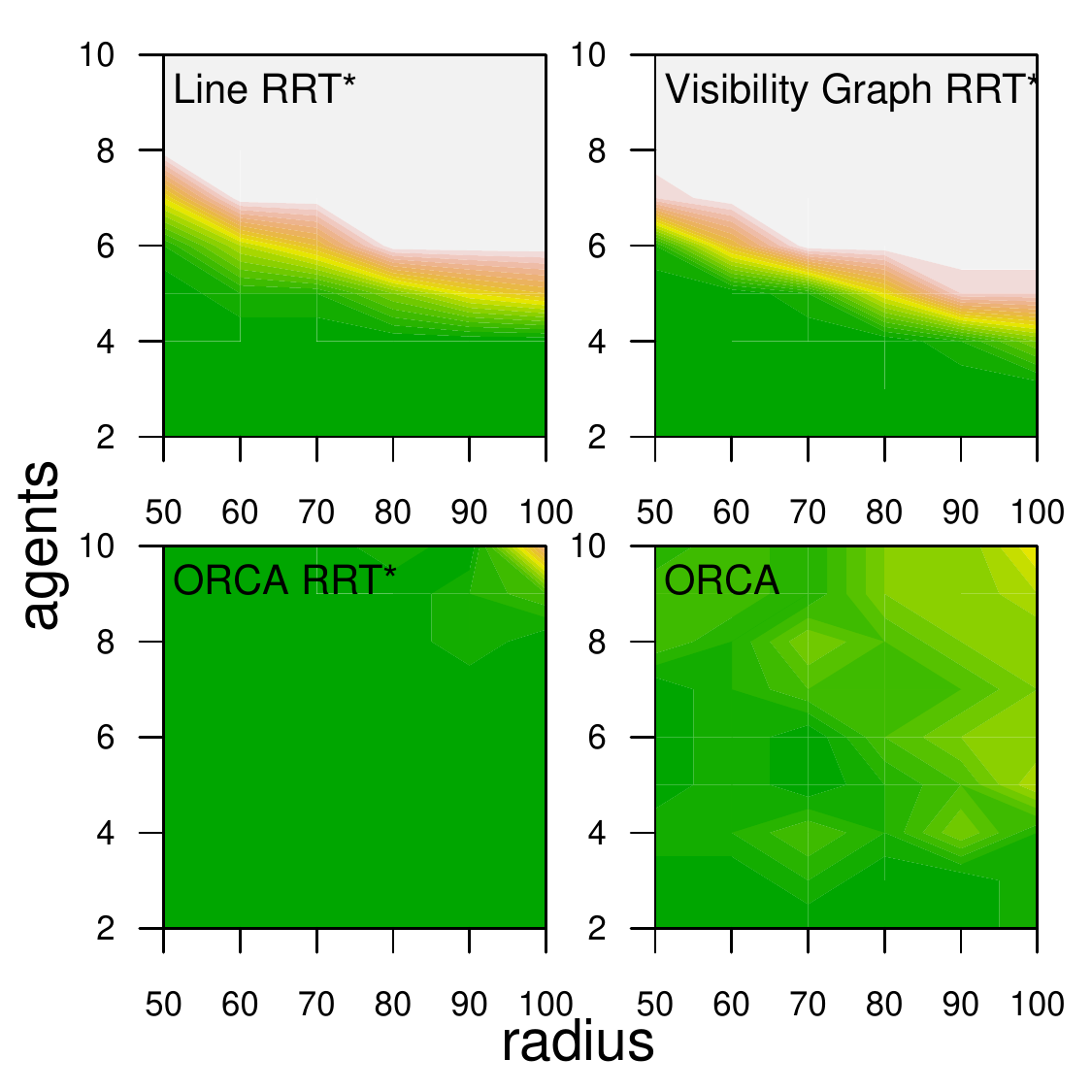}\label{figSuccessrateDoor}}
\subfigure[Cross environment]{\includegraphics[width=\imagewidth]{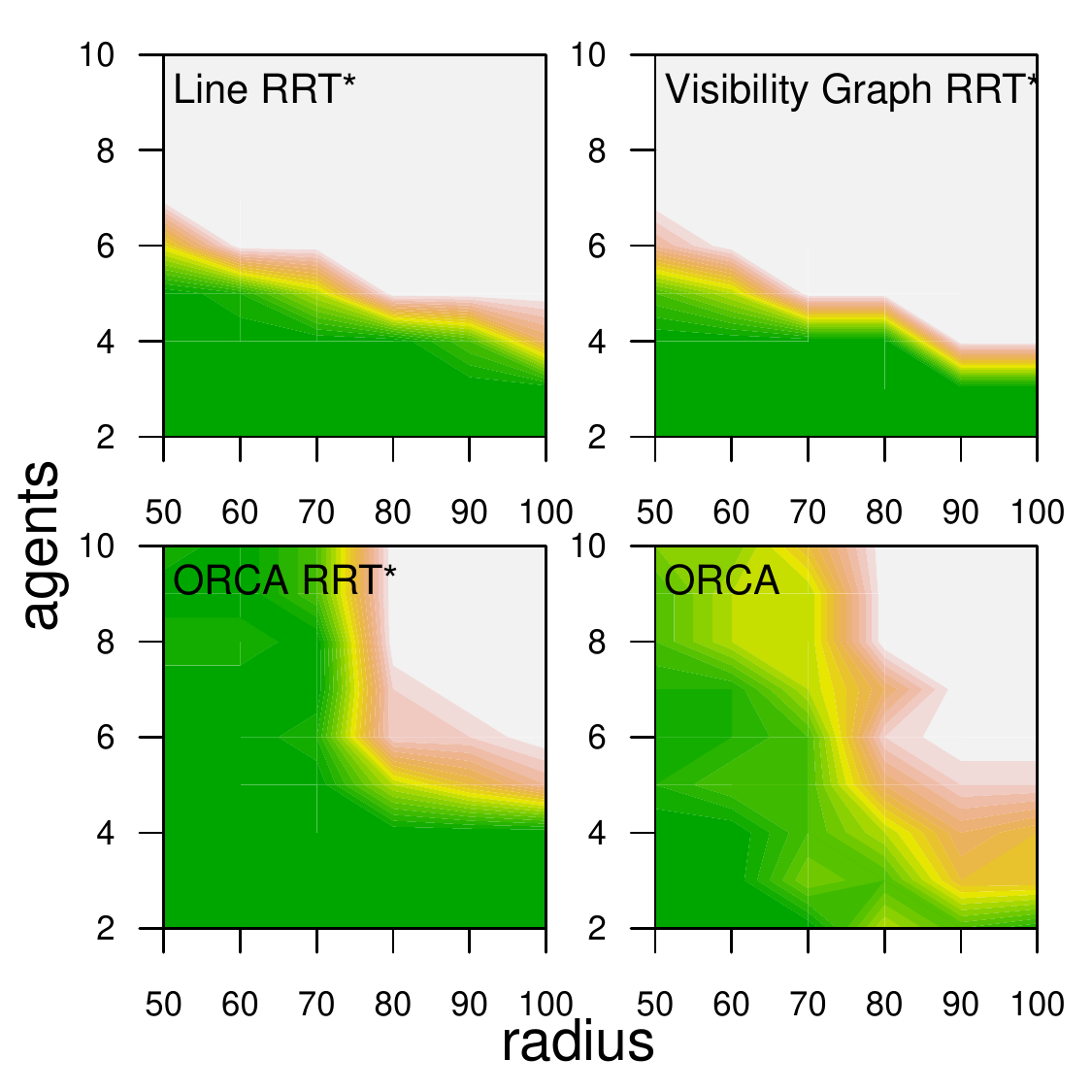}\label{figSuccessrateCross}}
\subfigure[Maze environment]{\includegraphics[width=\imagewidth]{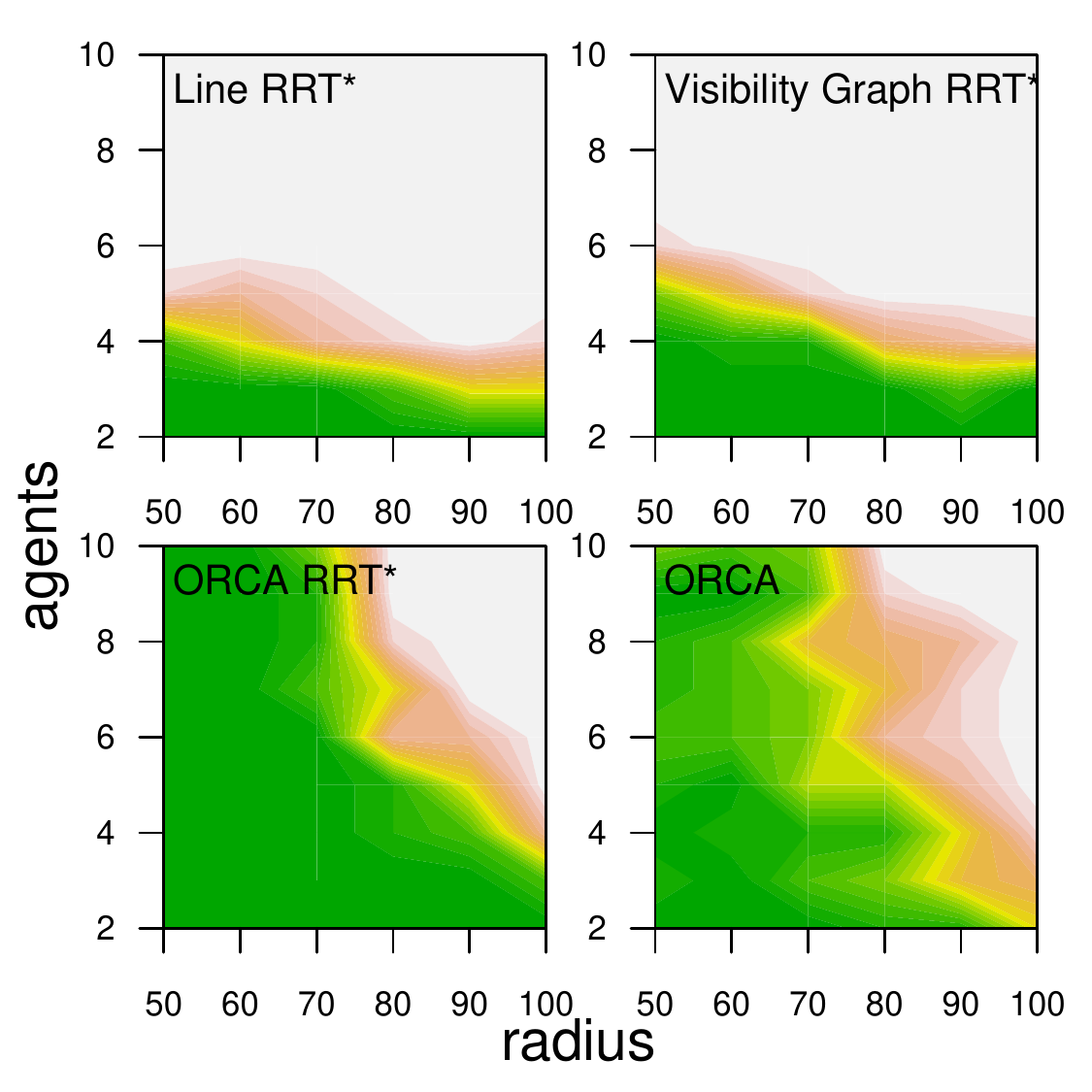}\label{figSuccessrateMazeInf}}
\subfigure{\includegraphics[height=\imagewidth]{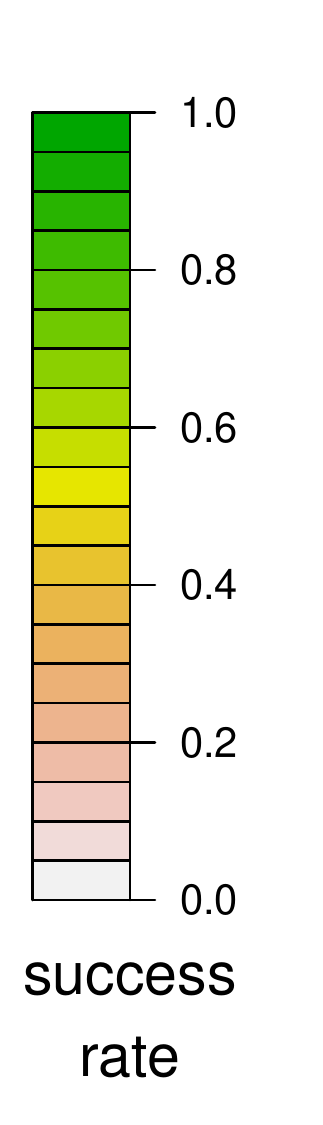}}
\vspace{-3mm}
\caption{Success rate of tested algorithms on test instances, no threshold}
\label{figSuccessrateInf}
\vspace{3mm}

\subfigure[Empty environment]{\includegraphics[width=\imagewidth]{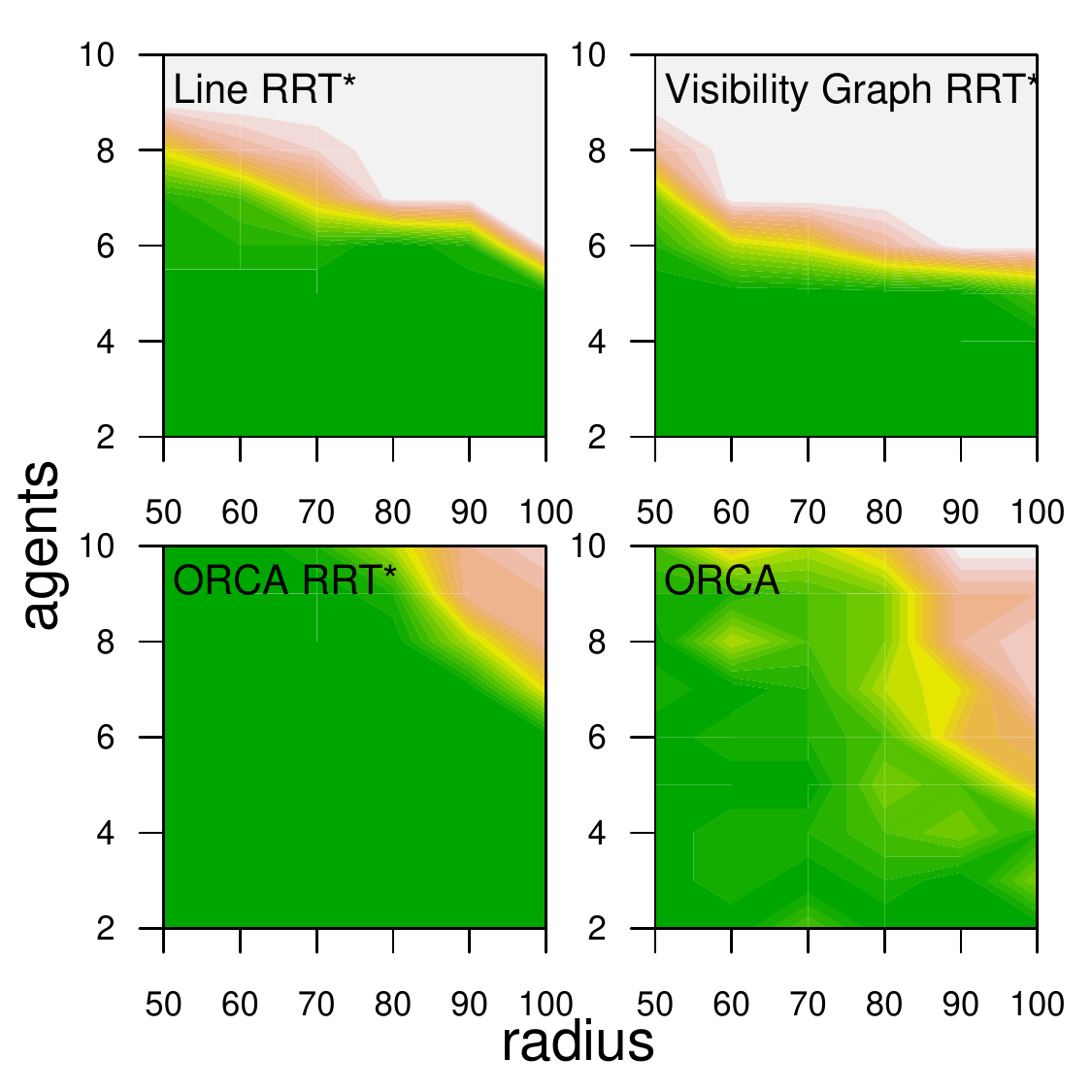}}
\subfigure[Door environment]{\includegraphics[width=\imagewidth]{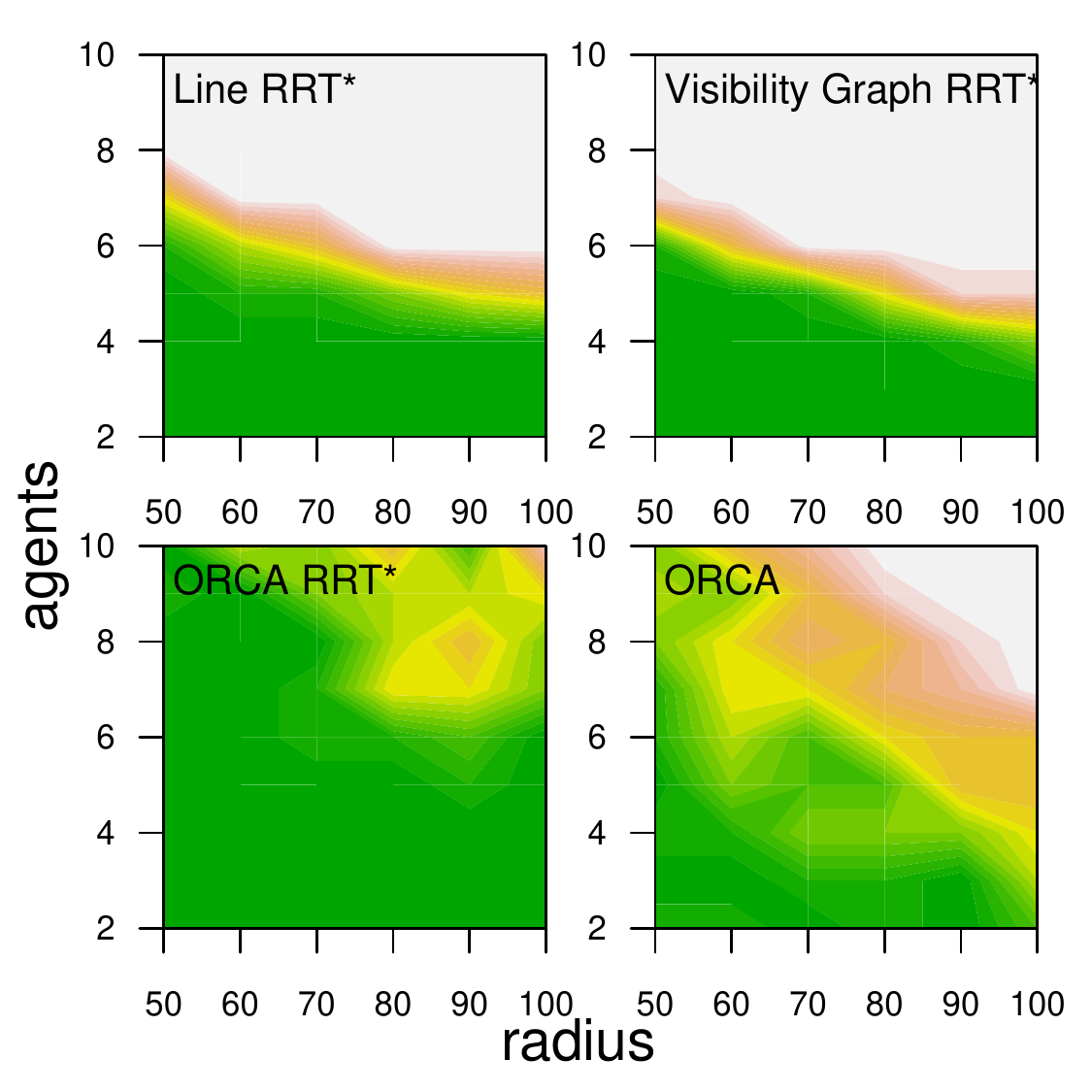}}
\subfigure[Cross environment]{\includegraphics[width=\imagewidth]{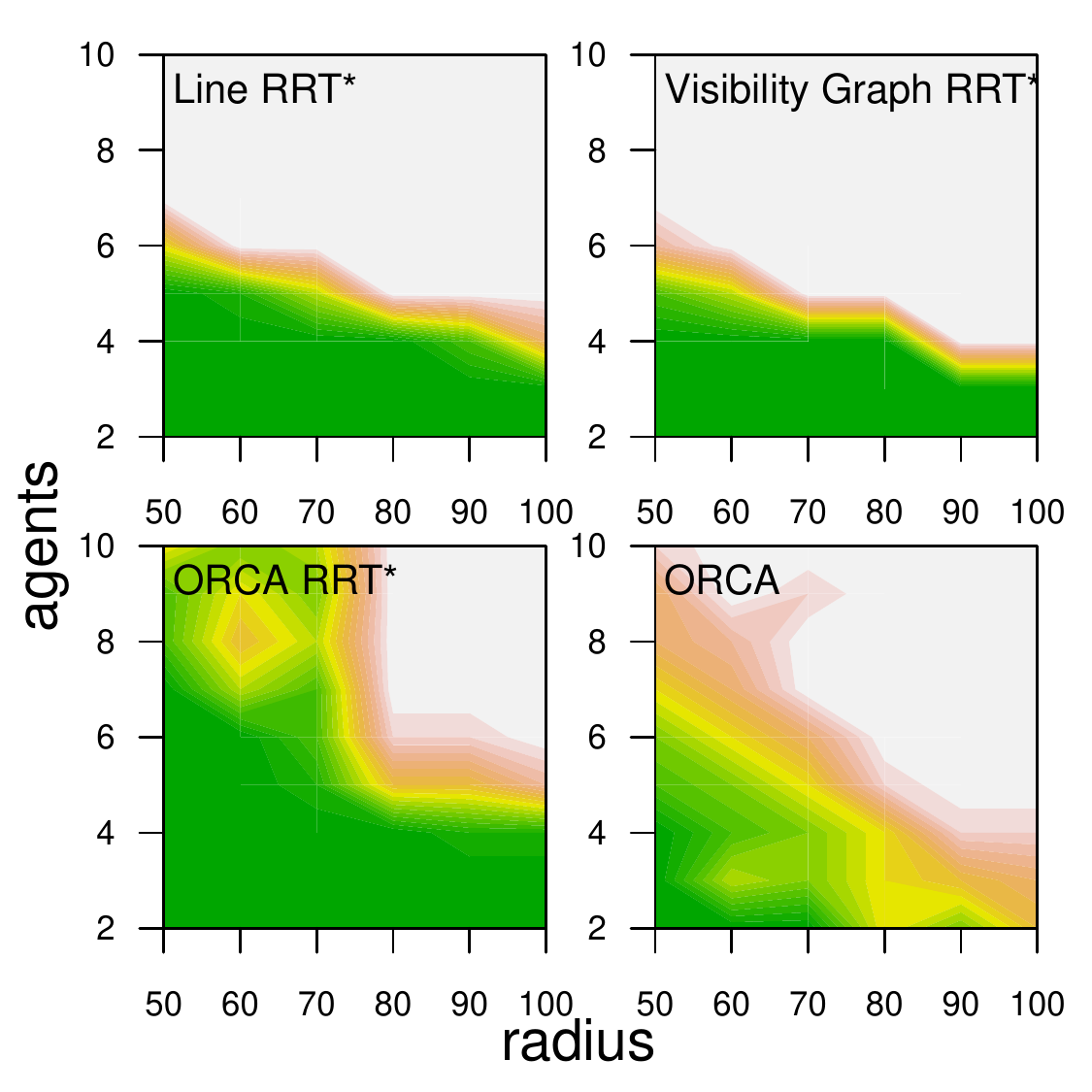}}
\subfigure[Maze environment]{\includegraphics[width=\imagewidth]{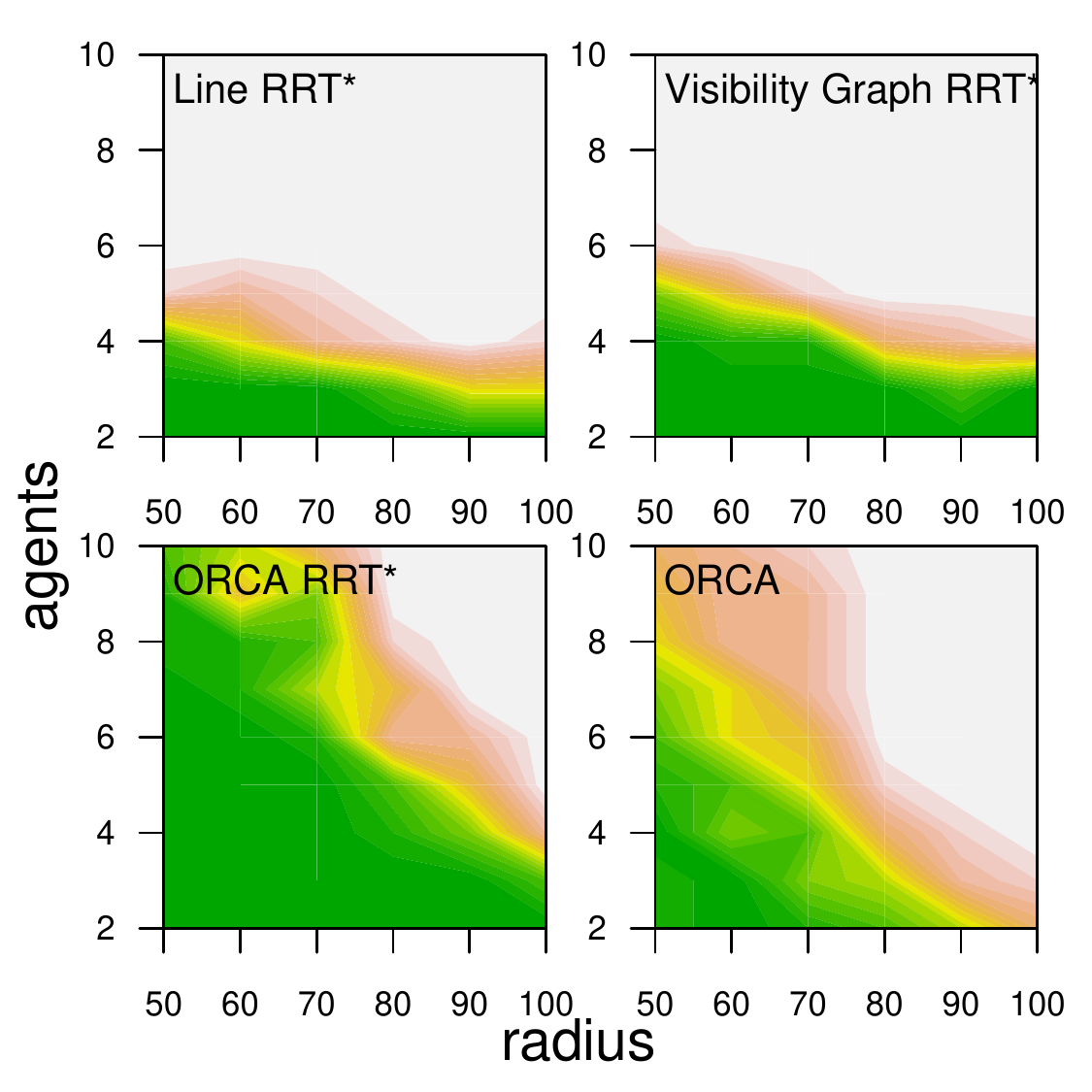}\label{figSuccessrateMaze5}}
\subfigure{\includegraphics[height=\imagewidth]{figures/success_radius_agents/legend.pdf}}
\vspace{-3mm}
\caption{Success rate of tested algorithms on test instances, threshold 5}
\label{figSuccessrate5}
\vspace{3mm}

\subfigure[Empty environment]{\includegraphics[width=\imagewidth]{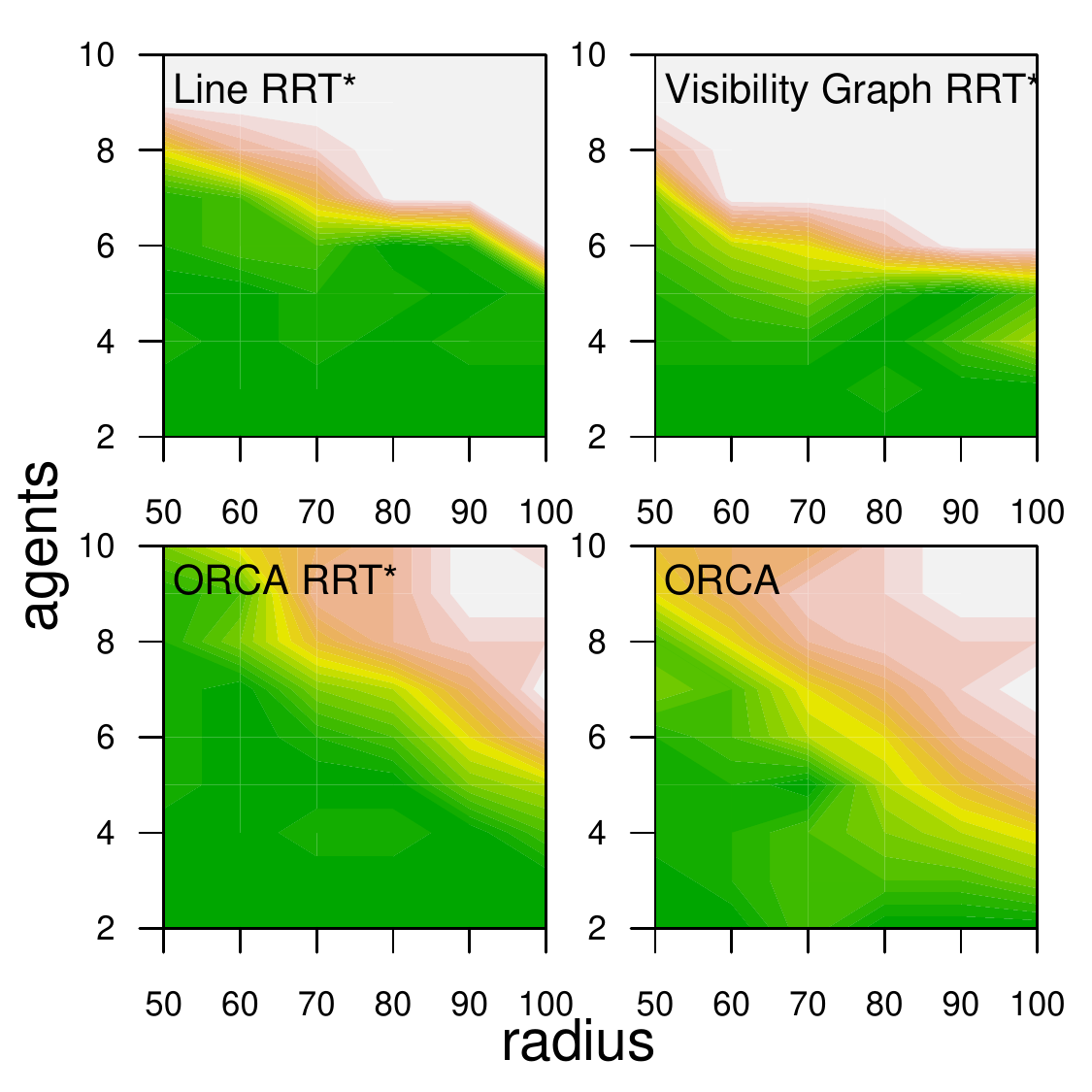}}
\subfigure[Door environment]{\includegraphics[width=\imagewidth]{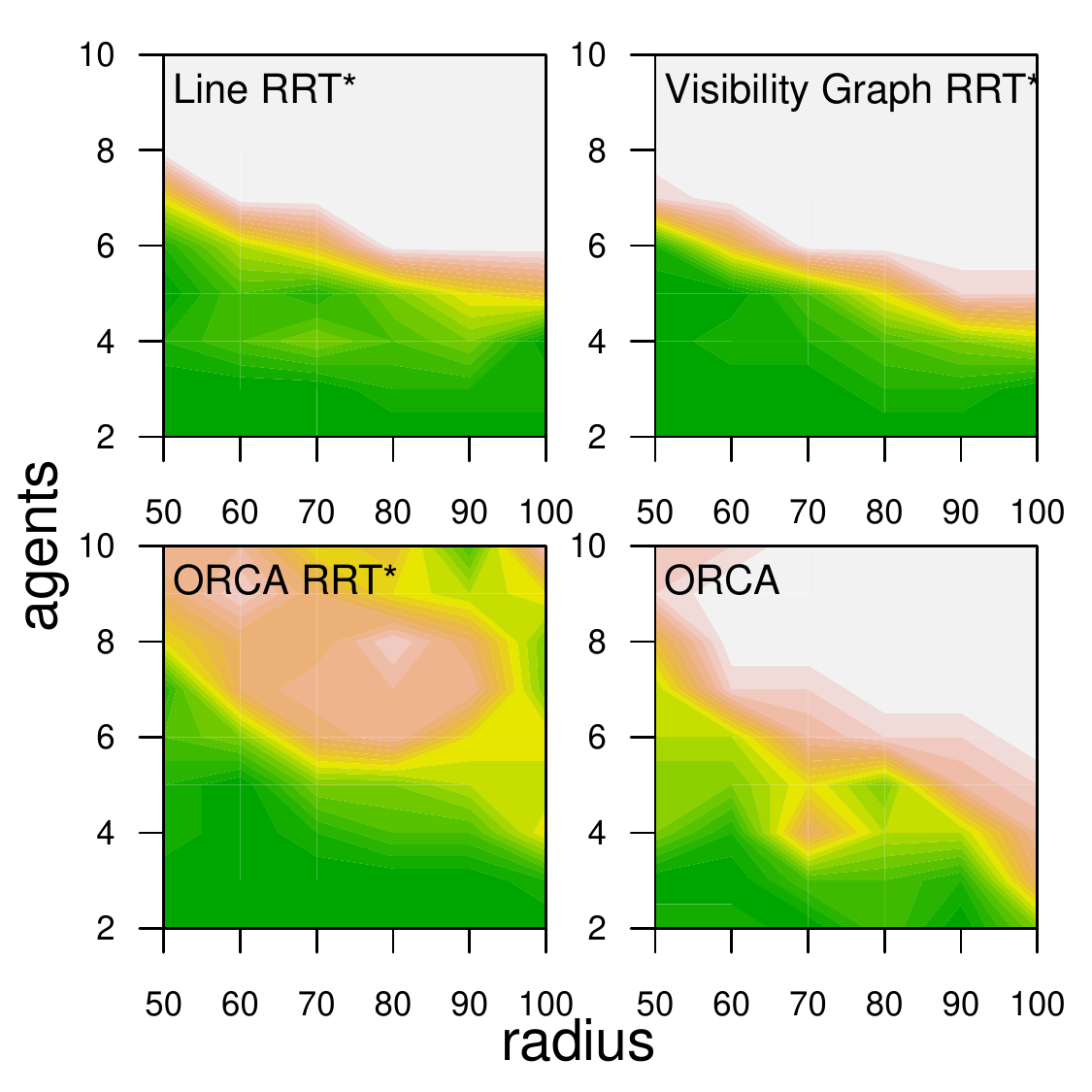}}
\subfigure[Cross environment]{\includegraphics[width=\imagewidth]{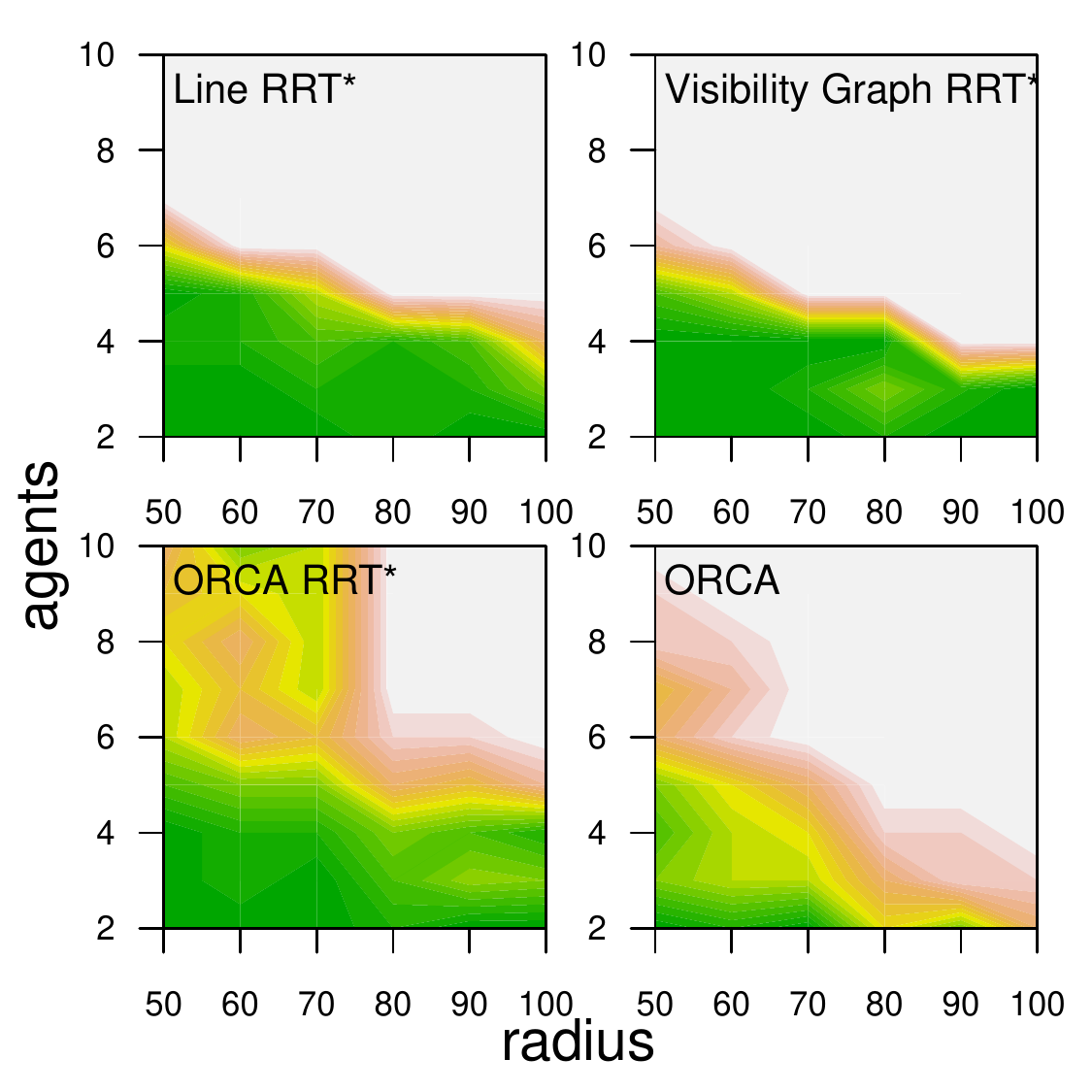}}
\subfigure[Maze environment]{\includegraphics[width=\imagewidth]{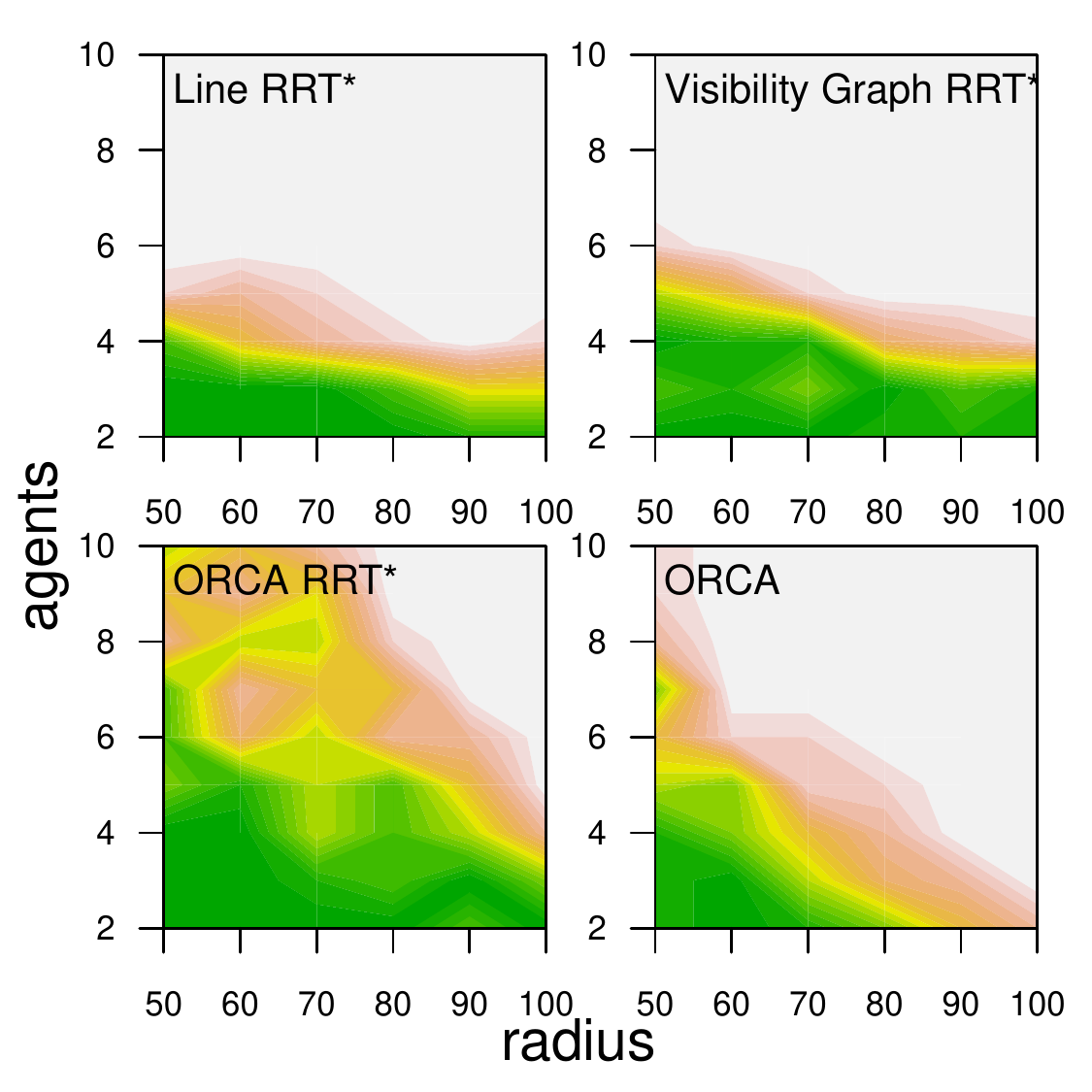}}
\subfigure{\includegraphics[height=\imagewidth]{figures/success_radius_agents/legend.pdf}}
\vspace{-3mm}
\caption{Success rate of tested algorithms on test instances, threshold 2.5}
\label{figSuccessrate2_5}
\centering
\subfigure[Empty environment]{\includegraphics[width=4.6cm]{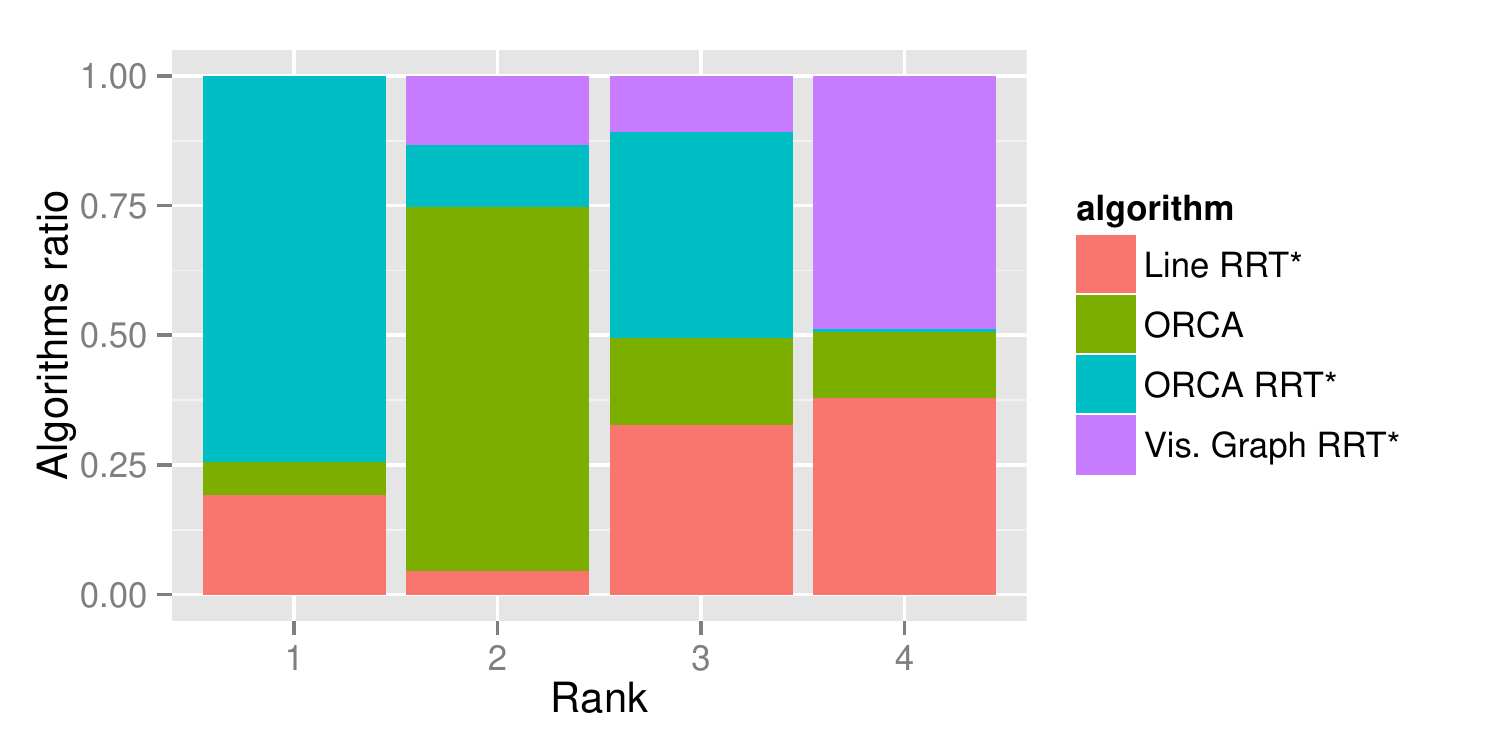}}
\hspace{-5mm}
\subfigure[Door environment]{\includegraphics[width=4.6cm]{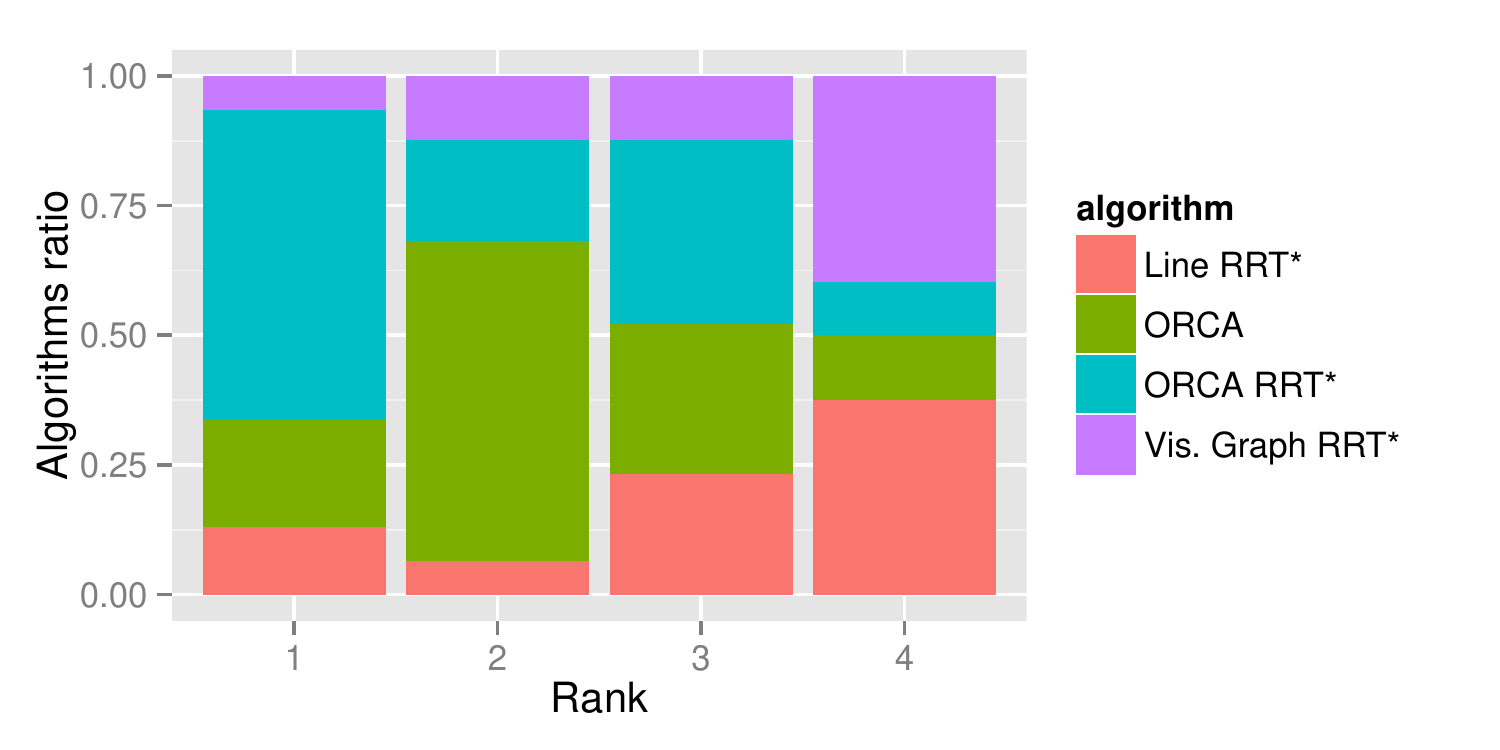}}
\hspace{-5mm}
\subfigure[Cross environment]{\includegraphics[width=4.6cm]{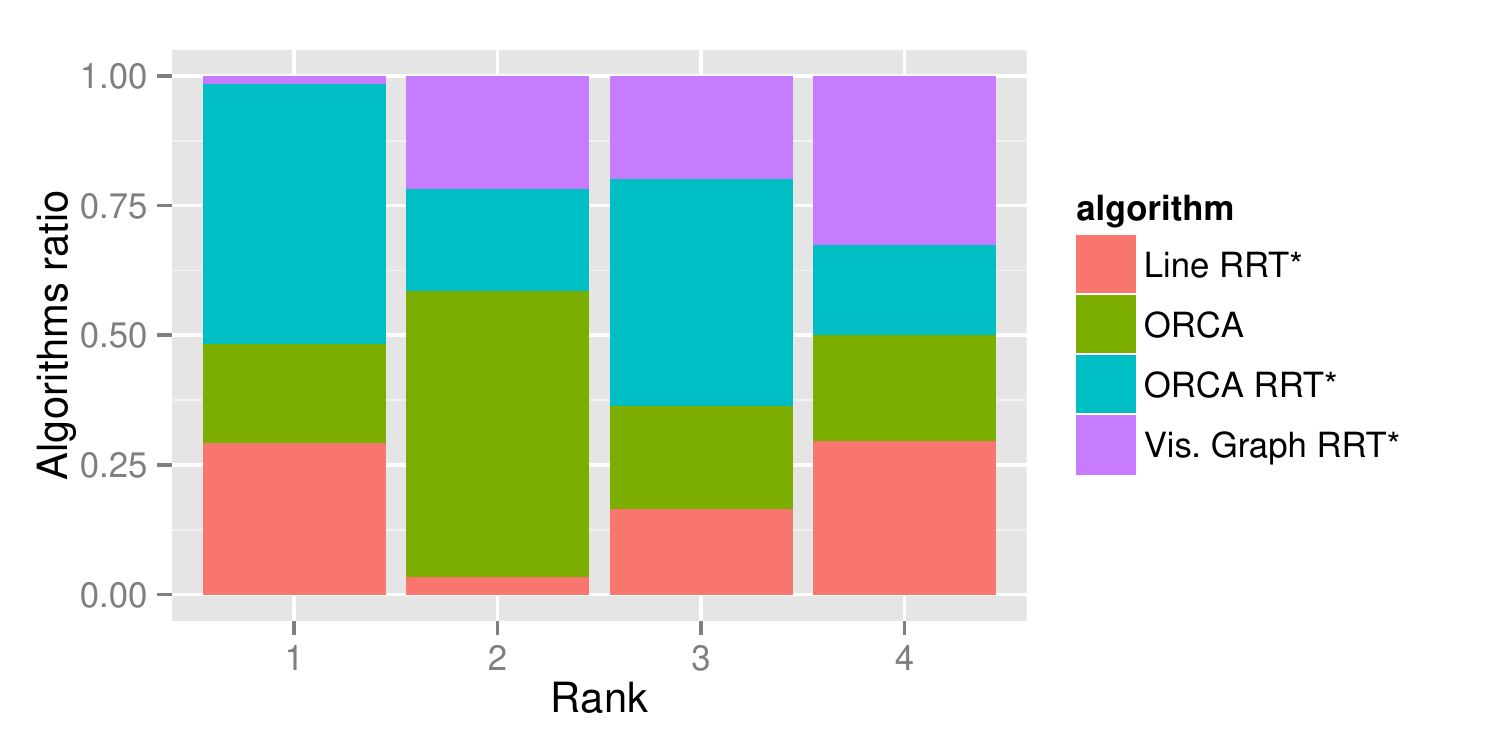}}
\hspace{-5mm}
\subfigure[Maze environment]{\includegraphics[width=4.6cm]{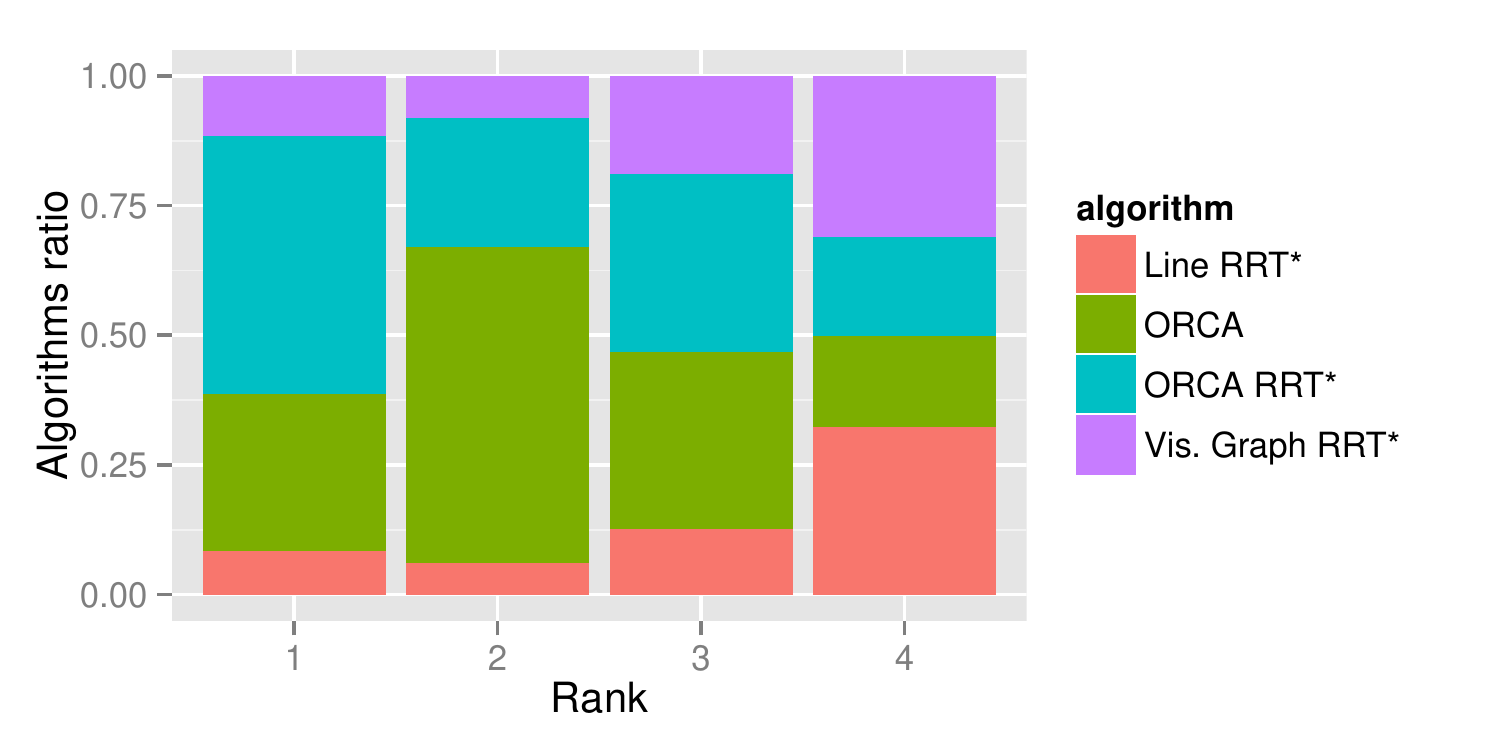}}
\vspace{-3mm}
\caption{Rank histograms for running time 5 seconds}
\label{figRank5}
\vspace{3mm}
\subfigure[Empty environment]{\includegraphics[width=4.6cm]{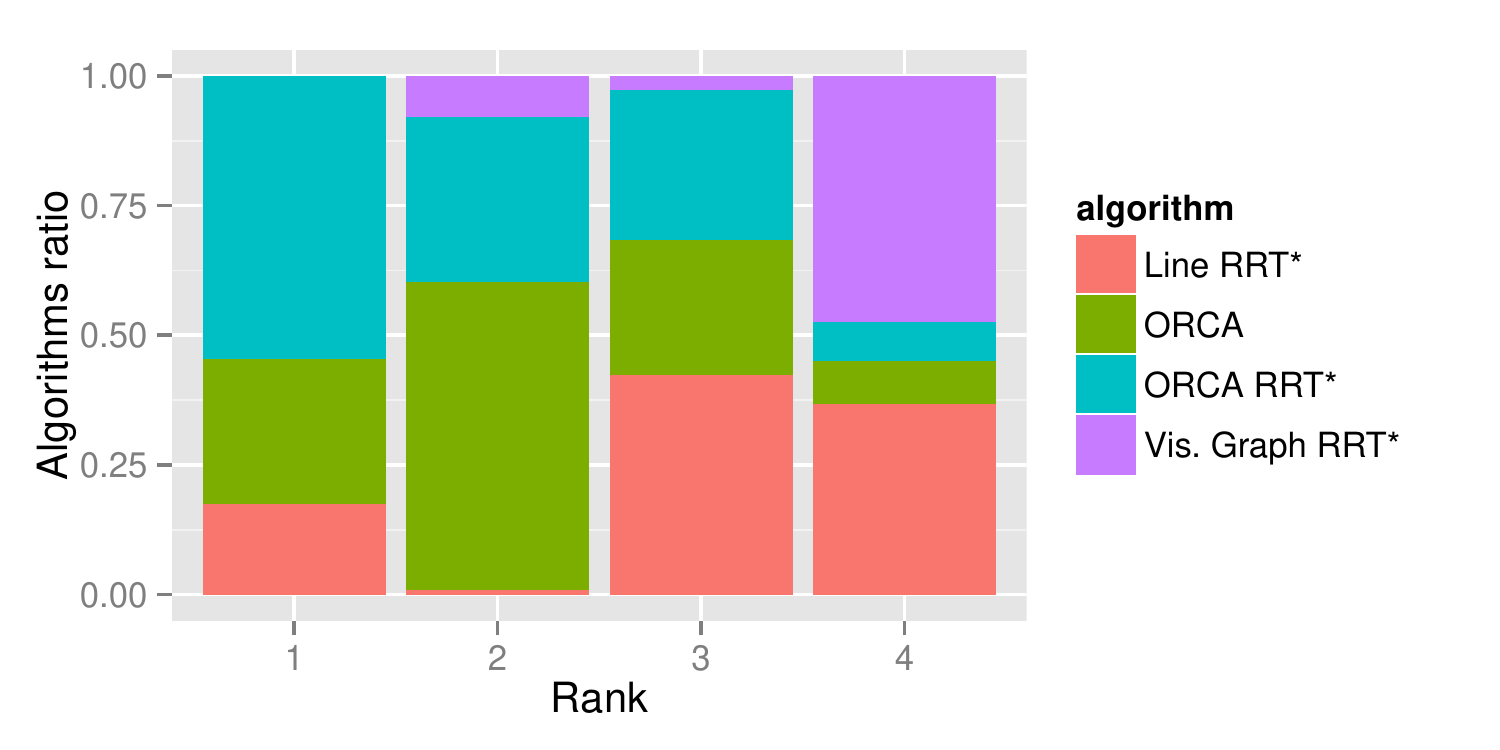}}
\hspace{-5mm}
\subfigure[Door environment]{\includegraphics[width=4.6cm]{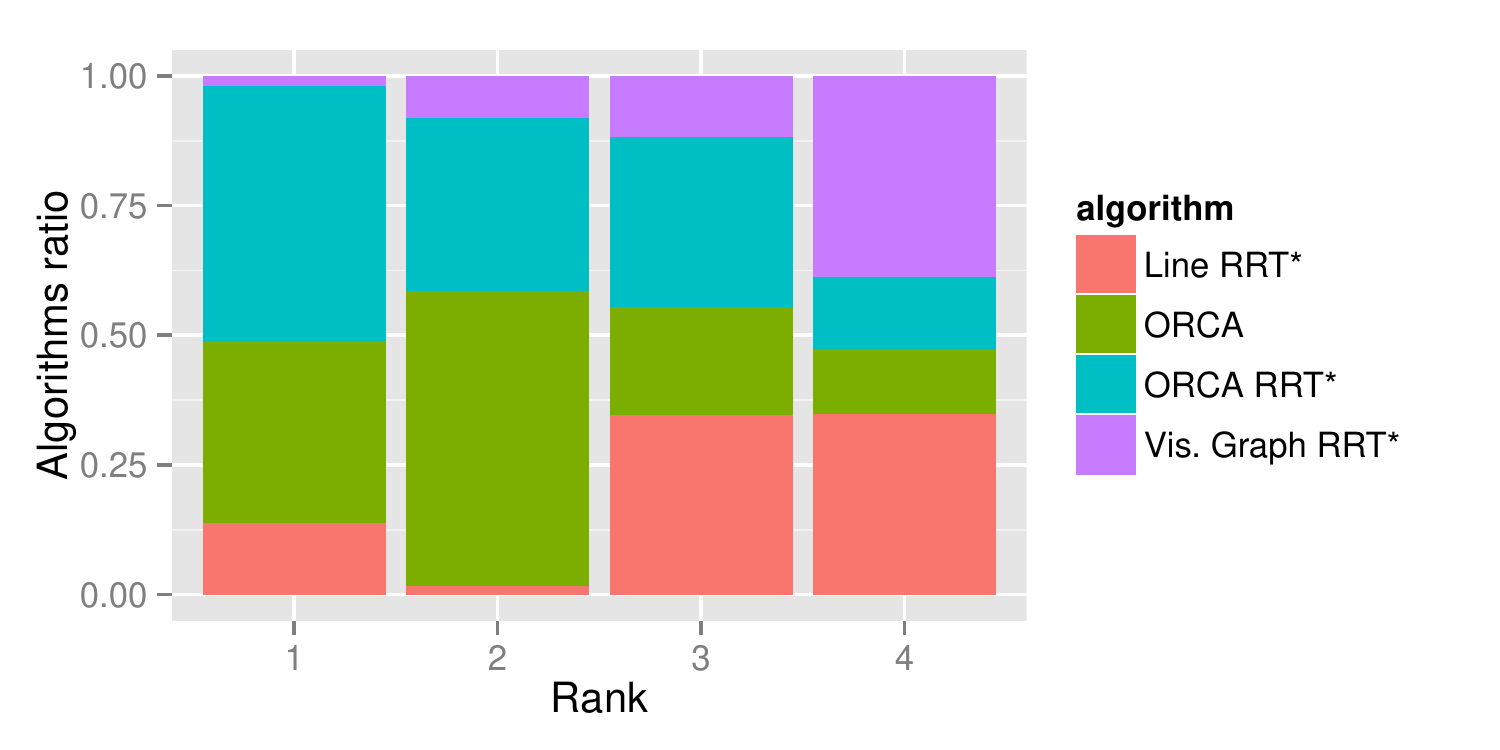}}
\hspace{-5mm}
\subfigure[Cross environment]{\includegraphics[width=4.6cm]{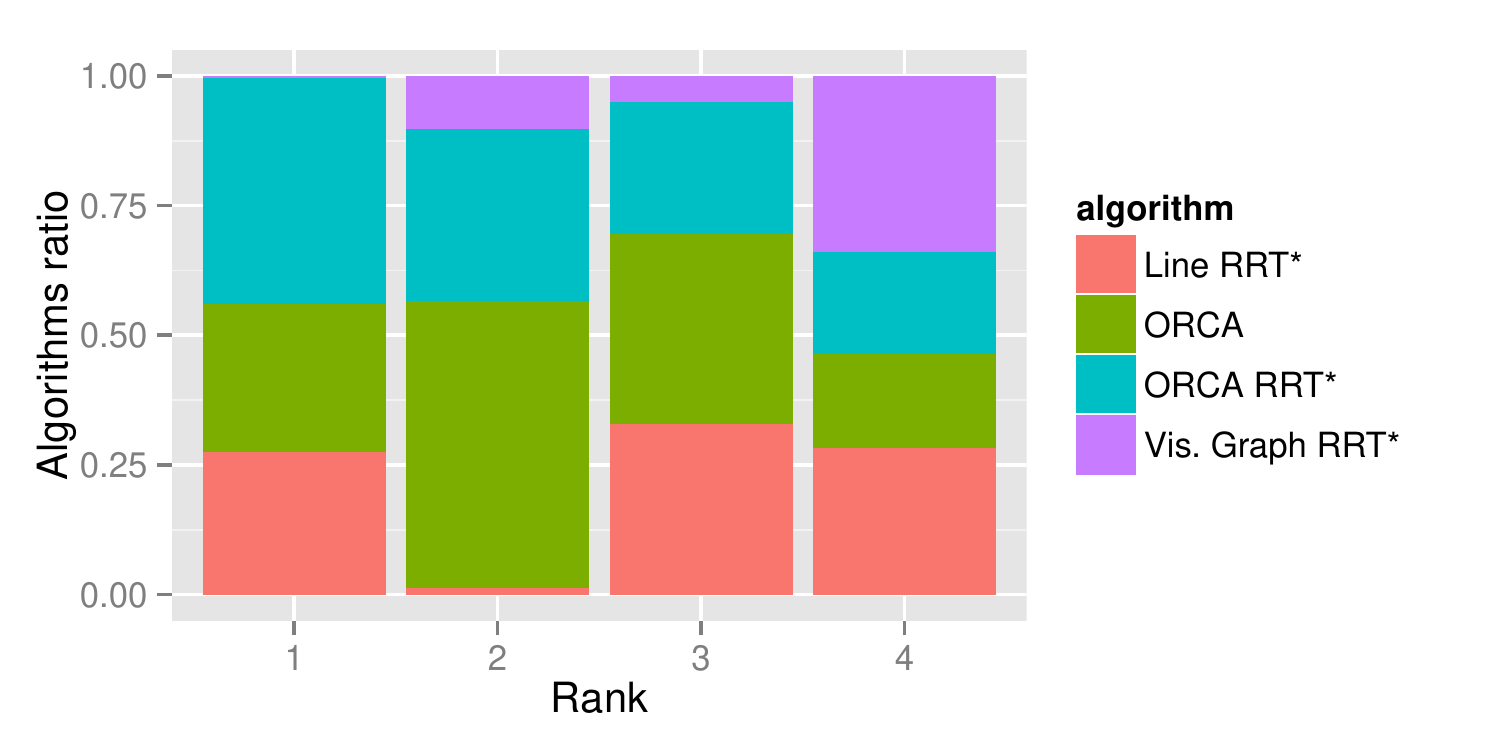}}
\hspace{-5mm}
\subfigure[Maze environment]{\includegraphics[width=4.6cm]{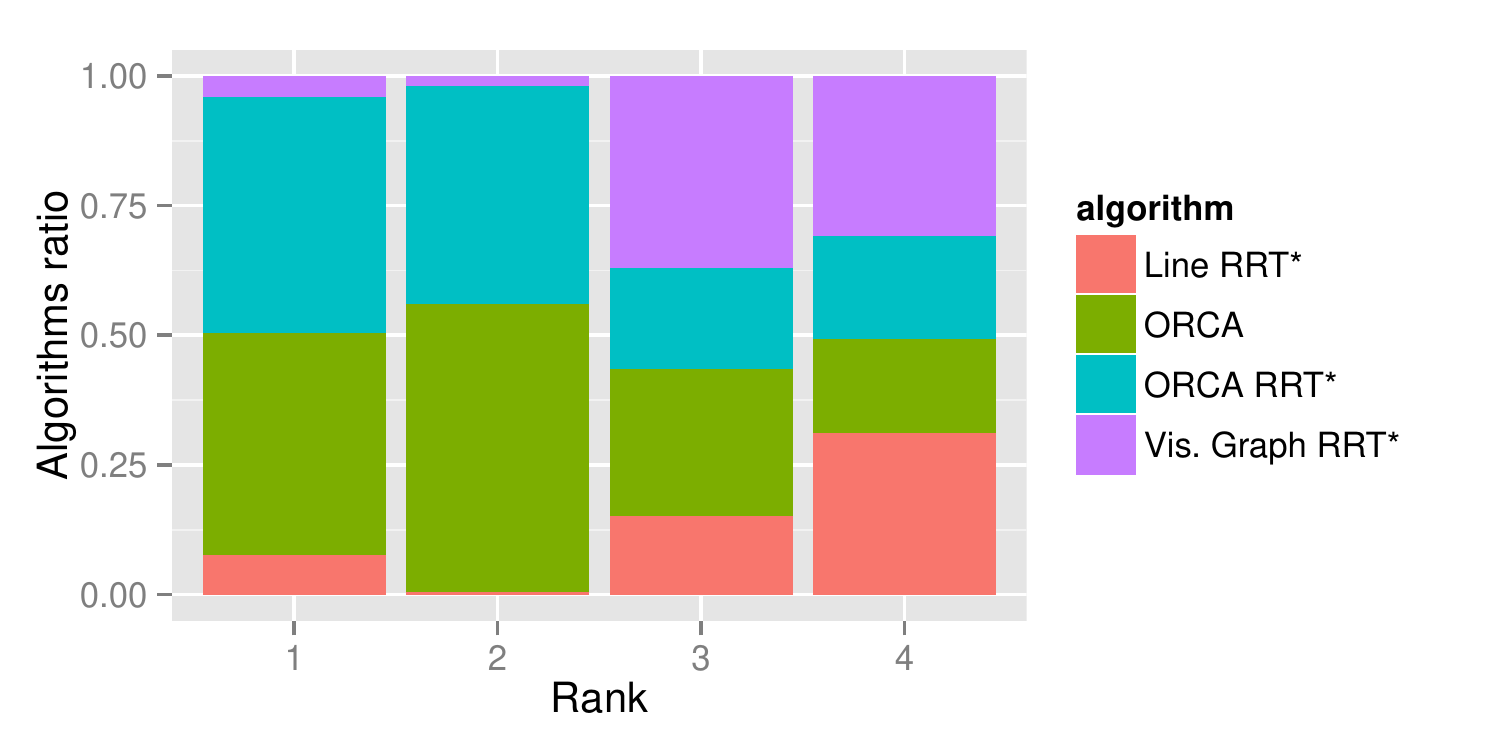}}
\vspace{-3mm}
\caption{Rank histograms for running time 1 second}
\label{figRank1}
\end{figure*}
%

\newpage
\vspace*{-0.95cm}
\subsection{Success rate}\label{sec:SR}
We compare the success rate of four implemented algorithms with respect to different numbers of agents and different radii of agents. The limit on runtime of the tested algorithms has been set to 5 seconds. To illustrate the distribution of solution quality, we plot these graphs for different suboptimality thresholds:
\begin{itemize}
\item Figure~\ref{figSuccessrateInf} uses no suboptimality threshold and contains all instances.
\item In Figure~\ref{figSuccessrate5} we consider all solutions having suboptimality over 5 as unsuccessful.
\item In Figure~\ref{figSuccessrate2_5} we consider all solutions having suboptimality over 2.5 as unsuccessful.
\end{itemize}

We can observe three significant phenomena. Firstly, the success rate of RRT* based algorithms Line-RRT* and Visibility Graph-RRT* drops fast with increasing number of agents. This behavior was expected because the planning takes place in a state space exponential in the number of agents. These algorithms are therefore able to solve the defined problems for the number of agents up to approximately 5 in the studied scenarios. Note that RRT* is provably probabilistically complete~\cite{karaman}, but for some instances given the limited runtime the algorithm was not able to find even the first feasible solution. On the other hand the success rate of both RRT* algorithms is stable with changing threshold.

Second, the success rate of the ORCA reactive technique drops with the increasing radii of the agents. This can be partly explained by the existence of corridors in the test scenarios that with the increasing radii of the agents become "narrow" and therefore hard to solve locally.  If the agents are small, then they can swap positions without leaving the corridor, which is easy to achieve using local collision avoidance. If the agents are large, then there will be corridors that can accommodate only one agent at a time, which requires longer term planning, where one agent keeps the way clear for the other agent to pass which is only achievable with planning approaches.

Further, we can observe that the solutions generated by ORCA are often of low quality (notice the difference between Figure~\ref{figSuccessrateMazeInf} and~\ref{figSuccessrateMaze5}). This typically happens in crowded environments, where are the agents likely to get stuck in slowly evolving deadlock situations.

Third, the success rate of the ORCA-RRT* algorithm is close to one for both high number of agents and high agent radius. It drops only with the combination of high extremes of both parameters. This behavior is achieved by the combination of planning and reactive approaches. The planning component is able to solve the instances that require planning, while the reactive component is able to solve instances with higher numbers of agents.

Moreover, we can see that the success rate of ORCA-RRT* deteriorates much less rapidly with decreasing threshold than pure ORCA, which implies that ORCA-RRT* in general finds in the given runtime limit higher quality solutions than pure ORCA alone. Since the first extension in ORCA-RRT* is in fact identical to running pure ORCA, these results confirm that the algorithm exhibits incremental behavior, i.e. it improves the quality of the generated solution in time.

An important observation is that the there are instances that neither of the algorithms was able to solve on its own, but that got solved when the two algorithms were combined. For many other instances the combination of RRT* and ORCA provided a higher-quality result within the given runtime limit than each of the algorithms alone.

\subsection{Suboptimality}
Figure~\ref{figRank5} and~\ref{figRank1} show the histograms of ranks assigned to algorithms for run-time limits 5 and 1 seconds. A rank from 1 to 4 is assigned to each algorithm for each experiment according to its solution suboptimality compared to other algorithms. If two algorithms achieve the same suboptimality, the ranks are assigned to them randomly. If an algorithm was not able to find any solution, its rank is 4. Rank 1 means that an algorithm achieved lowest suboptimality for particular problem instance. Difference between figures \ref{figRank5} and \ref{figRank1} shows how the ranks of the algorithms (e.g. solution quality -- suboptimality) depend on the running time limit.
We observe that VisibilityGraph-RRT* algorithm achieved the worst ranks. Line-RRT* is slightly better due to it's ability to sample the state space very fast. ORCA algorithm achieved the second rank and ORCA-RRT* always achieved the first rank in the majority of problem instances.
The difference between Figure~\ref{figRank5} and~\ref{figRank1} shows that while ORCA finds the best solution early, it does not benefit from added runtime. The variants of RRT* incrementally improve the solution and thus their performance is more dependent on the given runtime limit.

The results of the experiments confirm the ability of ORCA-RRT* to find higher-quality solutions compared to both RRT* variants. As a complement to the best problem instance set coverage, the ORCA-RRT* is also dominant in terms of the quality of the returned solution.

\section{Conclusions}

In this paper we studied the problem of finding coordinated paths for holonomic agents in 2-d polygonal environments. This problem is challenging due to its prohibitive complexity. We studied several RRT*-based algorithms for multi-agent coordinated path finding and  a reactive approach ORCA.
We found that while both approaches have limited coverage of the problem instance space, an approach combining planning and reactive technique benefits from both its parts, providing a better  problem instance set coverage and higher solution quality.

We call the new algorithm ORCA-RRT*. While RRT*-based algorithms often suffer from the exponential growth of the state space and thus are unable to solve instances with high number of agents, the reactive part of ORCA-RRT* is able to overcome this problem. On the other hand reactive techniques are often unable to solve problems containing local minima. Due to its RRT* planning part the ORCA-RRT* algorithm can avoid such local minima by random sampling of the state space.

ORCA-RRT* is an anytime algorithm, which can iteratively improve the provided solution. We experimented with several running time limits and examined the differences in the provided solutions.
In the future work we plan to deploy the ORCA-RRT* on hardware agents and investigate possibility of the extension towards non-holonomic agents.

\paragraph*{Acknowledgements}
The presented work was supported by the Czech Republic Ministry of Education, Youth and Sports, grants no. 7H11102 (D3CoS) and no. LD12044, and
by the ARTEMIS Joint Undertaking under the number 269336 (www.d3cos.eu).

\bibliographystyle{abbrv}
\bibliography{ORCA-RRT}  

\end{document}